\documentclass[10pt,journal]{IEEEtran}
\usepackage{multirow}
\usepackage{amsfonts}
\usepackage{amssymb}
\usepackage{amsmath}
\usepackage{subfigure}
\usepackage{graphicx}
\usepackage{mathrsfs}
\usepackage{algorithm}
\usepackage{algorithmic}
\usepackage{url}
\usepackage{threeparttable}
\usepackage{booktabs}

\newcommand{\x}{\mbox{\textit{x}}}
\newcommand{\vv}{\mbox{\textit{v}}}
\newcommand{\uu}{\mbox{\textit{u}}}
\newcommand{\bbeta}{\boldsymbol{\beta}}
\newcommand{\y}{\mbox{\textit{y}}}
\newcommand{\rr}{\mbox{\textit{r}}}
\newcommand{\dd}{\mbox{\textit{d}}}
\newcommand{\zz}{\mbox{\textit{z}}}
\newcommand{\s}{\mbox{\textit{s}}}

\def\theequation{\arabic{equation}}

\graphicspath{{./figures/}}

\begin{document}
\title{Bayesian Nonparametric Dictionary Learning for Compressed Sensing MRI}

\author{Yue Huang$^{\dagger}$, John Paisley$^{\dagger}$, Qin Lin, Xinghao Ding$^{\ddagger}$, Xueyang Fu and Xiao-ping Zhang, \IEEEmembership{Senior Member,~IEEE}

\thanks{This work supported by the National Natural Science Foundation of China (Nos. 30900328, 61172179, 61103121, 81301278), the Fundamental Research Funds for the Central Universities (Nos. 2011121051, 2013121023) and the Natural Science Foundation of Fujian Province of China (No. 2012J05160).\protect

Y. Huang, Q. Lin, X. Ding and X. Fu are with the Department of Communications Engineering at Xiamen University in Xiamen, Fujian, China. 

J. Paisley is with the Department of Electrical Engineering at Columbia University in New York, NY, USA. 

X.-P. Zhang is with the Department of Electrical and Computer Engineering at Ryerson University in Toronto, Canada.\protect

$^{\dagger}$Equal contributions. $^{\ddagger}$Corresponding author: dxh@xmu.edu.cn
}}

\maketitle

\begin{abstract}
We develop a Bayesian nonparametric model for reconstructing magnetic resonance images (MRI) from highly undersampled $k$-space data. We perform dictionary learning as part of the image reconstruction process. To this end, we use the beta process as a nonparametric dictionary learning prior for representing an image patch as a sparse combination of dictionary elements. The size of the dictionary and the patch-specific sparsity pattern are inferred from the data, in addition to other dictionary learning variables. Dictionary learning is performed directly on the compressed image, and so is tailored to the MRI being considered. In addition, we investigate a total variation penalty term in combination with the dictionary learning model, and show how the denoising property of dictionary learning removes dependence on regularization parameters in the noisy setting. We derive a stochastic optimization algorithm based on Markov Chain Monte Carlo (MCMC) for the Bayesian model, and use the alternating direction 
method of multipliers (ADMM) for efficiently performing total variation minimization. We present empirical results on several MRI, which show that the proposed regularization framework can improve reconstruction accuracy over other methods.
\end{abstract}

\begin{IEEEkeywords}
compressed sensing, magnetic resonance imaging, Bayesian nonparametrics, dictionary learning
\end{IEEEkeywords}

\section{Introduction}
Magnetic resonance imaging (MRI) is a widely used technique for visualizing the structure and functioning of the body. A limitation of MRI is its slow scan speed during data acquisition. Therefore, methods for accelerating the MRI process have been heavily researched. Recent advances in signal reconstruction from measurements sampled below the Nyquist rate, called compressed sensing (CS) \cite{ref1,ref2}, have had a major impact on MRI \cite{ref3}. CS-MRI allows for significant undersampling in the Fourier measurement domain of MR images (called $k$-space), while still outputting a high-quality image reconstruction. While image reconstruction using this undersampled data is a case of an ill-posed inverse problem, compressed sensing theory has shown that it is possible to reconstruct a signal from significantly fewer measurements than mandated by traditional Nyquist sampling if the signal is sparse in a particular transform domain.

Motivated by the need to find a sparse domain for representing the MR signal, a large body of literature now exists on reconstructing MRI from significantly undersampled $k$-space data. Existing improvements in CS-MRI mostly focus on ($i$) seeking sparse domains for the image, such as contourlets \cite{ref6,ref7}; ($ii$) using approximations of the $\ell_0$ norm for better reconstruction performance with fewer measurements, for example $\ell_1$, FOCUSS, $\ell_p$ quasi-norms with $0 < p < 1$, or using smooth functions to approximate the $\ell_0$ norm \cite{ref8,ref11}; and ($iii$) accelerating image reconstruction through more efficient optimization techniques \cite{ref12,ref14,ref36}. In this paper we present a modeling framework that is similarly motivated.

CS-MRI reconstruction algorithms tend to fall into two categories: Those which enforce sparsity directly within some image transform domain \cite{ref3}--\cite{ref12}, \cite{ref14,ref15,ref17}, and those which enforce sparsity in some underlying latent representation of the image, such as an adaptive dictionary-based representation \cite{ref13,ref19}. Most CS-MRI reconstruction algorithms belong to the first category. For example Sparse MRI \cite{ref3}, the leading study in CS-MRI, performs MR image reconstruction by enforcing sparsity in both the wavelet domain and the total variation (TV) of the reconstructed image. Algorithms with image-level sparsity constraints such as Sparse MRI typically employ an off-the-shelf basis, which can usually capture only one feature of the image. For example, wavelets recover point-like features, while contourlets recover curve-like features. Since MR images contain a variety of underlying features, such as edges and textures, using a basis not adapted to the image can be 
considered a drawback of these algorithms. 

Finding a sparse basis that is suited to the image at hand can benefit MR image reconstruction, since CS theory shows that the required number of measurements is linked to the sparsity of the signal in the selected transform domain. Using a standard basis not adapted to the image under consideration will likely not provide a representation that can compete in sparsity with an adapted basis. To this end, dictionary learning, which falls in the second group of algorithms, learns a sparse basis on image subregions called patches that is adapted to the image class of interest. Recent studies in the image processing literature have shown that dictionary learning is an effective means for finding a sparse, patch-level representation of an image \cite{ref24,ref25,ref34}. These algorithms learn a patch-level dictionary by exploiting structural similarities between patches extracted from images within a class of interest. Among these approaches, adaptive dictionary learning---where the dictionary is learned directly 
from the image being considered---based on patch-level sparsity constraints usually outperforms analytical dictionary approaches in denoising, super-resolution reconstruction, interpolation, inpainting, classification and other applications, since the adaptively learned dictionary suits the signal of interest \cite{ref24}--\cite{ref29}. 

Dictionary learning has previously been applied to CS-MRI to learn a sparse basis for reconstruction, e.g., \cite{ref19}. With these methods, parameters such as the dictionary size and patch sparsity are preset, and algorithms are considered that are non-Bayesian. In this paper, we consider a new dictionary learning algorithm for CS-MRI that is motivated by Bayesian nonparametric statistics. Specifically, we consider a nonparametric dictionary learning model called BPFA \cite{ref32} that uses the beta process to learn the sparse representation necessary for CS-MRI reconstruction. The beta process is an effective prior for nonparametric learning of latent factor models; in this case the latent factors correspond to dictionary elements. While the dictionary size is therefore infinite in principle, through posterior inference the beta process learns a suitably compact dictionary in which the signal can be sparsely represented.

We organize the paper as follows. In Section \ref{sec.background} we review CS-MRI inversion methods and the beta process for dictionary learning. In Section \ref{sec:proposed}, we describe the proposed regularization framework and algorithm. We derive a Markov Chain Monte Carlo (MCMC) sampling algorithm for stochastic optimization of the dictionary variables in the objective function. In addition, we consider including a sparse total variation (TV) penalty, for which we perform efficient optimization using the alternating direction method of multipliers (ADMM). We then show the advantages of the proposed Bayesian nonparametric regularization framework on several CS-MRI problems in Section \ref{sec.experiments}.

\section{Background and Related work}\label{sec.background}
We use the following notation: Let $\x\in\mathbb{C}^N$ be a $\sqrt{N}\times\sqrt{N}$ MR image in vectorized form. Let $\mathcal{F}_u\in\mathbb{C}^{u\times N}$, $u\ll N$, be the undersampled Fourier encoding matrix and $\y=\mathcal{F}_u \x \in \mathbb{C}^u$ represent the sub-sampled set of $k$-space measurements. The goal is to estimate $\x$ from the small fraction of $k$-space measurements $\y$. For dictionary learning, let $R_i$ be the $i^{th}$ patch extraction matrix. That is, $R_i$ is a $P\times N$ matrix of all zeros except for a one in each row that extracts a vectorized $\sqrt{P}\times\sqrt{P}$ patch from the image, $R_i\x \in \mathbb{C}^P$ for $i=1,\dots,N$. We use overlapping image patches with a shift of one pixel and allow a patch to wrap around the image at the boundaries for mathematical convenience \cite{ref20,ref29}. All norms are extended to complex vectors when necessary, $\|a\|_p = \left(\sum_i|a_i|^p\right)^{1/p}$, where $|a_i|$ is the modulus of the complex number $a_i$.

\subsection{Two approaches to CS-MRI inversion}

We focus on single-channel CS-MRI inversion via optimizing an unconstrained function of the form
\begin{equation}\label{eqn.objective}
\arg\min_x~h(\x)+\frac{\lambda}{2}\|\mathcal{F}_u \x - \y\|_2^2,
\end{equation}
where $\|\mathcal{F}_u \x - \y\|_2^2$ is a data fidelity term, $\lambda > 0$ is a parameter and $h(\x)$ is a regularization function that controls properties of the image we want to reconstruct. As discussed in the introduction, the function $h$ can take several forms, but tends to fall into one of two categories according to whether image-level or patch-level information is considered. We next review these two approaches.

\subsubsection{Image-level sparse regularization}
CS-MRI with an image-level, or global regularization function $h_g(\x)$ is one in which sparsity is enforced within a transform domain defined on the entire image. For example, in Sparse MRI \cite{ref3} the regularization function is
\begin{equation}
h_g(\x)=\|W \x\|_1+\mu\, TV(\x),
\end{equation}
where $W$ is the wavelet basis and $TV(\x)$ is the total variation (spatial finite differences) of the image. Regularizing with this function requires that the image be sparse in the wavelet domain, as measured by the $\ell_1$ norm of the wavelet coefficients $\|W \x\|_1$, which acts as a surrogate for $\ell_0$ \cite{ref1,ref2}. The total variation term enforces homogeneity within the image by encouraging neighboring pixels to have similar values while allowing for sudden high frequency jumps at edges. The parameter $\mu>0$ controls the trade-off between the two terms. A variety of other image-level regularization approaches have been proposed along these lines, e.g., \cite{ref6,ref7,ref11}.

\subsubsection{Patch-level sparse regularization}
An alternative to the image-level sparsity constraint $h_g(\x)$ is a patch-level, or local regularization function $h_l(\x)$, which enforces that patches (square sub-regions of the image) have a sparse representation according to a dictionary. One possible general form of such a regularization function is,
\begin{equation}
 h_l(\x) = \sum_{i=1}^N \frac{\gamma}{2}\|R_i\x - D\alpha_i\|_2^2 + f(\alpha_i,D),
\end{equation}
where the dictionary matrix is $D \in \mathbb{C}^{P\times K}$ and $\alpha_i$ is a $K$-dimensional vector in $\mathbb{R}^K$. An important difference between $h_l(\x)$ and $h_g(\x)$ is the additional function $f(\alpha_i,D)$. While image-level sparsity constraints fall within a predefined transform domain, such as the wavelet basis, the sparse dictionary domain can be unknown for patch-level regularization and learned from data. The function $f$ enforces sparsity by learning a $D$ for which $\alpha_i$ is sparse.\footnote{The dependence of $h_l(\x)$ on $\alpha$ and $D$ is implied in our notation.} For example, \cite{ref13} uses K-SVD to learn $D$ off-line, and then approximately optimize the objective function
\begin{equation}\label{eqn.ksvd_objective} 
\arg\min_{\alpha_{1:N}}~ \sum_{i=1}^N \|R_i\x - D\alpha_i\|_2^2 \quad \mbox{subject to}~ \|\alpha_i\|_0 \leq T, ~\forall i,
\end{equation}
using orthogonal matching pursuits (OMP) \cite{ref26}. 
In this case, the $L_0$ penalty on the additional parameters $\alpha_i$ make this a non-convex problem. Using this definition of $h_l(\x)$ in (\ref{eqn.objective}), a local optimal solution can be found by an alternating minimization procedure \cite{ref46}: First solve the least squares solution for $\x$ using the current values of $\alpha_i$ and $D$, and then update $\alpha_i$ and $D$, or only $\alpha_i$ if $D$ is learned off-line.


\subsection{Dictionary learning with beta process factor analysis}\label{sec.BPFA}
Typical dictionary learning approaches require a predefined dictionary size and, for each patch, the setting of either a sparsity level $T$, or an error threshold $\epsilon$ to determine how many dictionary elements are used. In both cases, if the settings do not agree with ground truth, the performance can significantly degrade. Instead, we consider a Bayesian nonparametric method called beta process factor analysis (BPFA) \cite{ref32}, which has been shown to successfully infer both of these values, as well as have competitive performance with algorithms in several application areas \cite{ref32}--\cite{ref34a}, and see \cite{ref47}--\cite{ref50} for related algorithms. The beta process is driven by an underlying Poisson process, and so it's properties as a Bayesian nonparametric prior are well understood \cite{ref35b}. Originally used for survival analysis in the statistics literature, its use for latent factor modeling has been significantly increasing within the machine learning community 
\cite{ref32}--\cite{ref34a}, \cite{ref35a}, \cite{ref47}--\cite{ref50}.

\begin{algorithm}[t]
\caption{Dictionary learning with BPFA}\label{alg.bpfa}
\begin{enumerate}
\item Construct a dictionary $D = [\dd_1,\dots,\dd_K]$, with \vspace{-3pt}$$\dd_k\sim N{(0,P^{-1} I_P)},\quad k = 1,\dots,K.$$
\item \vspace{-3pt}Draw a probability \vspace{-3pt}$\pi_k \in [0,1]$ for each $\dd_k$: $$\pi_k \sim Beta{(c\gamma/K,c(1-\gamma/K))},\quad k = 1,\dots,K.$$
\item \vspace{-3pt}Draw precision values for noise and each weight \vspace{-3pt}$$\gamma_\varepsilon\sim Gam{(g_0,h_0)},\quad \gamma_{s} \sim Gam{(e_0,f_0)}.$$
\item \vspace{-3pt}For the $i^{th}$ patch in $\x$:
  \begin{enumerate}
  \item Draw the vector $s_i\sim N{(0,\gamma_{s}^{-1}I_K)}.$
  \item Draw the binary vector $z_i$ with $z_{ik}\sim Bern(\pi_k).$
  \item Define $\alpha_i = s_i\circ z_i$ by an element-wise product.
  \item Sample noisy patch $R_i\x \sim N(D\alpha_i,\gamma_\varepsilon^{-1}I_P)$.
  \end{enumerate}
\item Construct the image $\x$ as the average of all $R_i\x$ that overlap on a given pixel.
\end{enumerate}
\end{algorithm}

\subsubsection{Generative model} We give the original hierarchical prior structure of the BPFA model in Algorithm \ref{alg.bpfa}, extending this to complex-valued dictionaries in Section \ref{sec.algorithm}. With this approach, the model constructs a dictionary matrix $D\in\mathbb{R}^{P\times K}$ ($\mathbb{C}^{P\times K}$ below) of i.i.d.\ random variables, and assigns probability $\pi_k$ to vector $\dd_k$. The parameters for these probabilities are set such that most of the $\pi_k$ are expected to be small, with a few large. In Algorithm \ref{alg.bpfa} we use an approximation to the beta process.\footnote{For a finite $c > 0$ and $\gamma > 0$, the random measure $H = \sum_{k=1}^K \pi_k\delta_{d_k}$ converges weakly to a beta process as $K\rightarrow\infty$ \cite{ref35b,ref32a}.} Under this parameterization, each patch $R_i \x$ extracted from 
the image $\x$ is modeled as a sparse weighted combination of the dictionary elements, as determined by the element-wise 
product of $z_i \in \{0,1\}^K$ with the Gaussian vector $s_i$. What makes the model nonparametric is that for many values of $k$, the values of $z_{ik}$ will equal zero for all 
$i$ since $\pi_k$ will be very small; the model learns the number of these unused dictionary elements and their index values from the data. Therefore, the value of $K$ should be set to a large number that is more than the expected size of the dictionary. It can be shown that, under the assumptions of this prior, in the limit $K\rightarrow\infty$, the number of dictionary elements used by a patch is Poisson$(\gamma)$ distributed and the total number of dictionary elements used by the data grows like $c\gamma\ln N$, where $N$ is the number of patches \cite{ref35a}. The parameters of the model include $c,\gamma,e_0,f_0,g_0,h_0$ and $K$; we discuss setting these values in Section \ref{sec.experiments}.

\subsubsection{Relationship to K-SVD}\label{sec.KSVD} Another widely used dictionary learning method is K-SVD \cite{ref25}. Though they are models for the same problem, BPFA and K-SVD have some significant differences that we briefly discuss. K-SVD learns the sparsity pattern of the coding vector $\alpha_i$ using the OMP algorithm \cite{ref26} for each $i$. Holding the sparsity pattern fixed, it then updates each dictionary element and dimension of $\alpha$ jointly by a rank one approximation to the residual. Unlike BPFA, it learns as many dictionary elements as are given to it, so $K$ should be set wisely. BPFA on the other hand automatically prunes unneeded elements, and updates the sparsity pattern by using the posterior distribution of a Bernoulli process, which is significantly different from OMP. It updates the weights and the dictionary from their Gaussian posteriors as well. Because of this probabilistic structure, we derive a sampling algorithm for these variables that takes advantage of 
marginalization, and 
naturally learns the auxiliary variables $\gamma_\varepsilon$ and $\gamma_{s}$.

\begin{table}
\centering
\caption{Peak signal-to-noise ratio (PSNR) for image denoised by BPFA. Compared with K-SVD using correct (Match) and incorrect (Mismatch) noise parameter.\label{tab.denoising}}
\begin{tabular}{|c|c|c|c|c|}
\hline
\multicolumn{1}{|c|}{\multirow{2}{*}{$\sigma$}}&\multicolumn{2}{|c}{K-SVD denoising (PSNR)}&\multicolumn{2}{|c|}{BPFA denoising (PSNR)}\\
\cline{2-5}
\multicolumn{1}{|c|}{}&\multicolumn{1}{|c|}{Match} & \multicolumn{1}{|c|}{Mismatch}&\multicolumn{1}{|c|}{Results} & Learned noise\\
\hline
20/255	&32.28	&28.94	&32.88	&20.43/255\\
\hline
25/255	&31.08	&28.60	&31.81	&25.46/255\\
\hline
30/255	&29.99	&28.35	&30.94	&30.47/255\\
\hline
\end{tabular}
\end{table}

\begin{figure}
\centering
\subfigure[Noisy image]{\includegraphics[width=.21\textwidth]{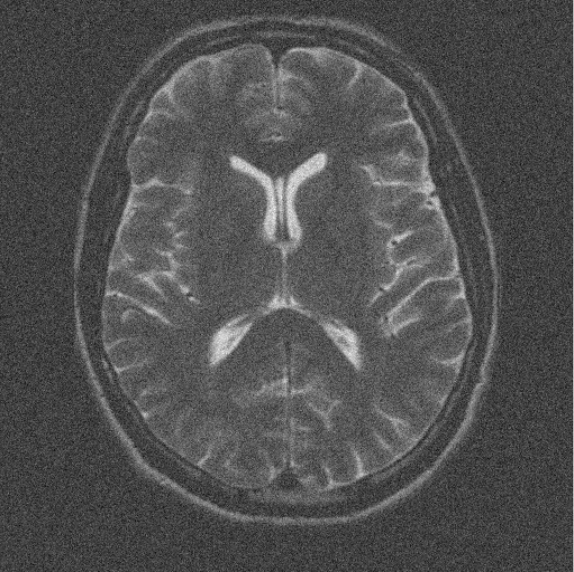}\label{fig:subfig1:a}}
\subfigure[Denoising by BPFA]{\includegraphics[width=.21\textwidth]{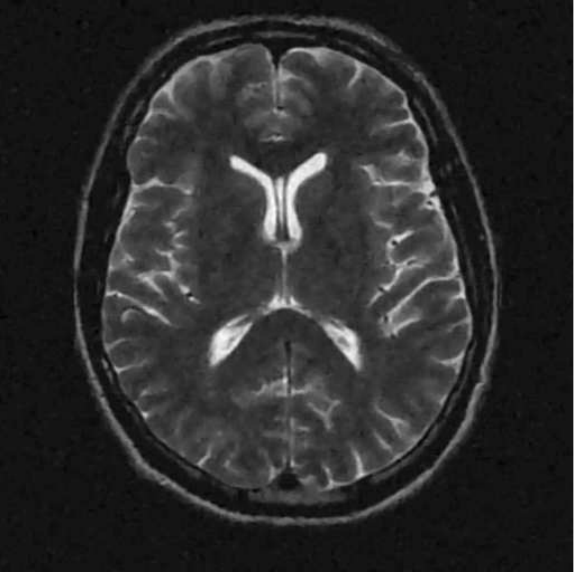}\label{fig:subfig2:b}}
\subfigure[Dictionary probabilities]{\includegraphics[width=.22\textwidth]{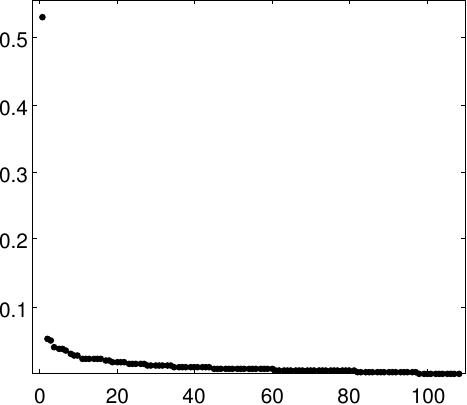}\label{fig:subfig2:c}}
\subfigure[Dictionary elements per patch]{\includegraphics[width=.22\textwidth]{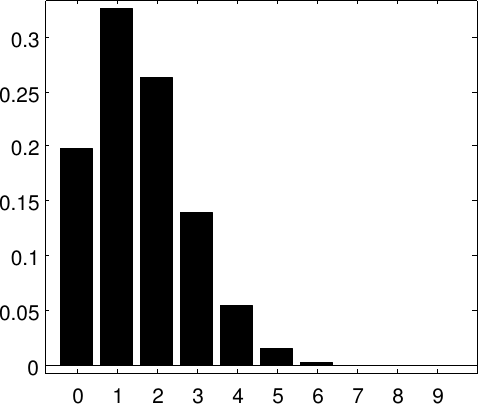}\label{fig:subfig2:d}}
\caption{(a)-(b) An example of denoising by BPFA (image scaled to [0,1]). (c) Shows the final probabilities of the dictionary elements and (d) shows a distribution on the number of dictionary elements used per patch.\label{fig:subfig2}}
\end{figure}

\subsubsection{Example denoising problem}\label{sec.denoising} As we will see, the relationship of dictionary learning to CS-MRI is essentially as a denoising step. To this end, we briefly illustrate BPFA on a denoising problem. Denoising of an image using dictionary learning proceeds by first learning the dictionary representation of each patch, $R_i\x \approx D \alpha_i$. The denoised reconstruction of $\x$ using BPFA is then $\x_{\mbox{\tiny{BPFA}}} = \frac{1}{P}\sum_i R_i^T D\alpha_i$.

We show an example using $6\times 6$ patches extracted from the noisy $512\times512$ image shown in Figure \ref{fig:subfig1:a}. In Figure \ref{fig:subfig2:b} we show the resulting denoised image. For this problem we truncated the dictionary size to $K=108$ and set all other model parameters to one. In Figures \ref{fig:subfig2:c} and \ref{fig:subfig2:d} we show some statistics from dictionary learning. For example, Figure \ref{fig:subfig2:c} shows the values of $\pi_k$ sorted, where we see that fewer than 100 elements are used by the data, many of which are very sparsely used. Figure \ref{fig:subfig2:d} shows the empirical distribution of the number of elements used per patch. We see the ability of the model to adapt the sparsity to the complexity of the patch. In Table \ref{tab.denoising} we show PSNR results for three noise variance levels. For K-SVD, we consider the case when the error parameter matches the ground truth, and when it mismatches it by a magnitude of five. As expected, when K-SVD does not 
have an appropriate parameter setting the performance suffers. BPFA on the other hand adaptively infers this value, which helps improve the denoising.

\section{CS-MRI with BPFA and TV penalty}\label{sec:proposed}
We next present our approach for reconstructing single-channel MR images from highly undersampled $k$-space data. In reference to the discussion in Section \ref{sec.background}, we consider a sparsity constraint of the form
\begin{equation}\label{eqn.our_objective}
\arg\min_{x,\varphi}~ \lambda_g h_g{(\x)}+h_l{(\x)}+\frac{\lambda}{2}\|\mathcal{F}_u \x-\y\|_2^2,
\end{equation}
\begin{displaymath}
 h_g(\x) := TV(\x),\quad h_l(\x) := \sum_{i=1}^N \frac{\gamma_\varepsilon}{2}\|R_i\x - D\alpha_i\|_2^2 + f(\varphi_i).
\end{displaymath}
For the local regularization function $h_l(\x)$ we use BPFA as given in Algorithm \ref{alg.bpfa} in Section \ref{sec.BPFA}. The parameters to be optimized for this penalty are contained in the set $\varphi_i = \{D,\s_i,\zz_i,\gamma_{\varepsilon},\gamma_s,\pi\}$, and are defined in Algorithm \ref{alg.bpfa}. We note that only $\s_i$ and $\zz_i$ vary in $i$, while the rest are shared by all patches.  The regularization term $\gamma_\varepsilon$ is a model variable that corresponds to an inverse variance parameter of the multivariate Gaussian likelihood. This likelihood is equivalently viewed as the squared error penalty term in $h_l(\x)$ in (\ref{eqn.our_objective}).  This term acts as the sparse basis for the image and also aids in producing a denoised reconstruction, as discussed in Sections \ref{sec.BPFA}, \ref{sec.lambda} and \ref{sec.simulated}. For the global regularization function $h_g(\x)$ we use the total variation of the image. This term encourages homogeneity within contiguous regions of the image, 
while still allowing for sharp jumps in pixel value at edges due 
to the underlying $\ell_1$ penalty. The regularization parameters $\lambda_g$, $\gamma_\varepsilon$ and $\lambda$ control the trade-off between the terms in this optimization. Since we sample a new value of $\gamma_\varepsilon$ with each iteration of the algorithm discussed shortly, this trade-off is adaptively changing.

For the total variation penalty $TV(\x)$ we use the isotropic TV model. Let $\psi_i$ be the $2 \times N$ difference operator for pixel $i$. Each row of $\psi_i$ contains a $1$ centered on pixel $i$, and the first row also has a $-1$ on the pixel directly above pixel $i$, while the second has a $-1$ corresponding to the pixel to the right, and zeros elsewhere. Let $\Psi = [\psi_1^T,\dots,\psi_N^T]^T$ be the resulting $2N\times N$ difference matrix for the entire image. The TV coefficients are $\beta = \Psi\x \in \mathbb{C}^{2N}$, and the isotropic TV penalty is $TV(\x) = \sum_{i}\|\psi_i\x\|_2 = \sum_i \sqrt{|\beta|_{2i-1}^2 + |\beta|_{2i}^2}$, where $i$ ranges over the pixels in the MR image. For optimization we use the alternating direction method of multipliers (ADMM) \cite{ref45,ref44}. ADMM works by performing dual ascent on the augmented Lagrangian objective function introduced for the total variation coefficients. For completeness, we give a brief review of ADMM in the appendix.

\subsection{Algorithm}\label{sec.algorithm}
We present an algorithm for finding a local optimal solution to the non-convex objective function given in (\ref{eqn.our_objective}). We can write this objective as
$$L(\x,\varphi) =  \lambda_g\textstyle\sum_i\|\psi_i\x\|_2 +\textstyle\sum_i \frac{\gamma_\varepsilon}{2}\|R_i\x - D\alpha_i\|_2^2\quad$$
\begin{equation}
 + \textstyle\sum_i f(\varphi_i)+\frac{\lambda}{2}\|\mathcal{F}_u \x-\y\|_2^2.
\end{equation}
We seek to minimize this function with respect to $\x$ and the dictionary learning variables $\varphi_i = \{D,\s_i,\zz_i,\gamma_{\varepsilon},\gamma_s,\pi\}$.

Our first step is to put the objective into a more suitable form. We begin by defining the TV coefficients for the $i^{th}$ pixel as $\bbeta_i := [\beta_{2i-1}~ \beta_{2i}]^T = \psi_i\x$. We introduce the vector of Lagrange multipliers $\eta_i$, and then split $\bbeta_i$ from $\psi_i\x$ by relaxing the equality via an augmented Lagrangian. This results in the objective function 
\begin{eqnarray}\label{eqn.objective2}
L(\x,\beta,\eta,\varphi) \hspace{-8pt}&=&\hspace{-8pt} \sum_{i=1}^N \lambda_g\|\bbeta_i\|_2 + \eta_i^T(\psi_i\x-\bbeta_i ) + \frac{\rho}{2}\|\psi_i\x-\bbeta_i \|_2^2\nonumber\\
				  & & +~\sum_{i=1}^N \frac{\gamma_\varepsilon}{2}\|R_i\x - D\alpha_i\|_2^2 + f(\varphi_i)\nonumber\\
				  & &+~\frac{\lambda}{2}\|\mathcal{F}_u \x-\y\|_2^2.
\end{eqnarray}
From the ADMM theory \cite{ref46}, this objective will have (local) optimal values $\bbeta_i^*$ and $\x^*$ with $\bbeta^*_i = \psi_i\x^*$, and so the equality constraints will be satisfied \cite{ref45}.\footnote{For a fixed $D, \alpha_{1:N}$ and $\x$ the solution is also globally optimal.} Optimizing this function can be split into three separate sub-problems: one for TV, one for BPFA and one for updating the reconstruction $\x$. Following the discussion of ADMM in the appendix, we define $\uu_i = (1/\rho)\eta_i$ and complete the square in the first line of (\ref{eqn.objective2}). We then cycle through the following three sub-problems,\vspace{4pt}
\begin{equation}
\begin{array}{lrcl}
 (P1)  & \bbeta_i' \hspace{-4pt}& = &\hspace{-4pt} \arg\min_{\bbeta} ~ \lambda_g \|\bbeta\|_2 + \frac{\rho}{2}\|\psi_i\x -\bbeta  + \uu_i\|_2^2, \vspace{4pt}\nonumber\\\vspace{4pt}
 & \uu_i' \hspace{-4pt}&=&\hspace{-4pt} \uu_i + \psi_i\x - \bbeta'_i , \quad i=1,\dots,N, \vspace{4pt}\nonumber\\\vspace{4pt}
 (P2)  & \varphi'  \hspace{-4pt}& = &\hspace{-4pt} \arg\min_\varphi ~ \sum_i \frac{\gamma_\varepsilon}{2}\|R_i\x - D\alpha_i\|_2^2 + f(\varphi_i) ,\vspace{4pt}\nonumber\\
 (P3)  & \x' \hspace{-4pt}& = &\hspace{-4pt} \arg\min_x ~ \sum_i \frac{\rho}{2}\|\psi_i\x - \bbeta'_i  + \uu_i'\|_2^2\vspace{4pt} \nonumber\\
			    &&& ~~~+ \sum_i \frac{\gamma_\varepsilon'}{2}\|R_i\x - D'\alpha_i'\|_2^2 + \frac{\lambda}{2}\|\mathcal{F}_u \x-\y\|_2^2.\vspace{4pt}\nonumber\\
\end{array}
\end{equation}

Solutions for sub-problems $P1$ and $P3$ are globally optimal (conditioned on the most recent values of all other parameters). We cannot solve $P2$ analytically since the optimal values for the set of all BPFA variables do not have a closed form solution. Our approach for $P2$ is to use stochastic optimization by Gibbs sampling each variable of BPFA conditioned on current values of all other variables. We next present the updates for each sub-problem. We give an outline in Algorithm \ref{alg.basic}.\newline

\begin{algorithm}[t]
   \caption{Outline of algorithm}
   \label{alg.basic}
\raggedright\vspace{4pt}
\begin{description}
\item \hspace{-40pt} Input: $\y$ -- Undersampled $k$-space data\\
\item \hspace{-40pt} Output: $\x$ -- Reconstructed MR image\\
\item \hspace{-40pt} Initialize: $\x = \mathcal{F}_u^H \y$ and each $\uu_i = 0$. Sample $D$ from prior.\\\vspace{2pt}
\item\hspace{-30pt}{Step 1.} $P1$: Optimize each $\bbeta_i$ via shrinkage.\\
\item\hspace{-30pt}{Step 2.} Update Lagrange multiplier vectors $\uu_i$.\\
\item\hspace{-30pt}{Step 3.} $P2$: Gibbs sample BPFA variables once.\\
\item\hspace{-30pt}{Step 4.} $P3$: Solve for $\x$ using Fourier domain.\\
\item \textbf{if} \textit{not converged} \textbf{then} \textit{return to Step 1.}
\end{description}
\end{algorithm}

\subsubsection{Algorithm for P1 (total variation)}
We can solve for $\bbeta_i$ exactly for each pixel $i = 1,\dots,N$ by using a generalized shrinkage operation \cite{ref45},
\begin{equation}
 \bbeta_i' = \max\left\lbrace \|\psi_i\x+\uu_i\|_2 - \frac{\lambda_g}{\rho},0\right\rbrace \cdot \frac{\psi_i\x+\uu_i}{\|\psi_i\x+\uu_i\|_2}.
\end{equation}
We recall that $\bbeta_i$ corresponds to the 2-dimensional TV coefficients for pixel $i$, with differences in one direction vertically and horizontally. We then update the corresponding Lagrange multiplier, $\uu_i' = \uu_i + \psi_i\x - \bbeta_i'$.\newline

\subsubsection{Algorithm for P2 (BPFA)}
We update the parameters of BPFA using Gibbs sampling. We are therefore stochastically optimizing (\ref{eqn.objective2}), but only for this sub-problem. With reference to Algorithm \ref{alg.bpfa}, the P2 sub-problem entails sampling new values for the complex dictionary $D$, the binary vectors $\zz_i$ and real-valued weights $\s_i$ (with which we construct $\alpha_i = \s_i \circ \zz_i$ through the element-wise product), the precisions $\gamma_\varepsilon$ and $\gamma_{s}$, and the probabilities $\pi_{1:K}$, with $\pi_k$ giving the probability that $z_{ik} = 1$. In principle, there is no limit to the number of samples that can be made, with the final sample giving the updates used in the other sub-problems. We found that a single sample is sufficient in practice and leads to a faster algorithm. We describe the sampling procedure below.

\paragraph{Sample dictionary $D$} 
We define the $P\times N$ matrix $X = [R_1\x,\dots,R_N\x]$, which is a complex matrix of all vectorized patches extracted from the image $\x$. We also define the $K\times N$ matrix $\boldsymbol{\alpha} = [\alpha_1,\dots,\alpha_N]$ containing the dictionary weight coefficients for the corresponding columns in $X$ such that $D\boldsymbol{\alpha}$ is an approximation of $X$ to which we add noise from a circularly-symmetric complex normal distribution. The update for the dictionary $D$ is
\begin{eqnarray}\label{eqn.dict_up}
 D &=& X\boldsymbol{\alpha}^T(\boldsymbol{\alpha}\boldsymbol{\alpha}^T + (P/\gamma_\varepsilon)I_P)^{-1} + E,\\
 E_{p,:} &\stackrel{ind}{\sim}& \mathcal{CN}(0,(\gamma_\varepsilon\boldsymbol{\alpha}\boldsymbol{\alpha}^T + PI_P)^{-1}),\quad p = 1,\dots,P,\nonumber
\end{eqnarray}
where $E_{p,:}$ is the $p^{th}$ row of $E$. To sample this, we can first draw $E_{p,:}$ from a multivariate Gaussian distribution with this covariance structure, followed by an i.i.d.\ uniform rotation of each value in the complex plane. We note that the first term in Equation (\ref{eqn.dict_up}) is the $\ell_2$-regularized least squares solution for $D$. The addition of correlated Gaussian noise in the complex plane generates the sample from the conditional posterior of $D$. Since both the number of pixels and $\gamma_\varepsilon$ will tend to be very large, the variance of the noise is small and the mean term dominates the update for $D$.

\paragraph{Sample sparse coding $\alpha_i$} Sampling $\alpha_i$ entails sampling $s_{ik}$ and $z_{ik}$ for each $k$. We sample these values using block sampling. We recall that to block sample two variables from their joint conditional posterior distribution, $(s,z) \sim p(s,z|-)$, one can first sample $z$ from the marginal distribution, $z \sim p(z|-)$, and then sample $s|z \sim p(s|z,-)$ from the conditional distribution. The other sampling direction is possible as well, but for our problem sampling $z\rightarrow s|z$ is more efficient for finding a mode of the objective function. 

We define $\rr_{i,-k}$ to be the residual error in approximating the $i^{th}$ patch with the current values from BPFA minus the $k^{th}$ dictionary element, $\rr_{i,-k} = R_i\x - \sum_{j\neq k} (s_{ij}z_{ij})\dd_j $. We then sample $z_{ik}$ from its conditional posterior Bernoulli distribution $z_{ik} \sim p_{ik}\delta_1 + q_{ik}\delta_0$, where following a simplification,
\begin{eqnarray}
p_{ik} &\propto & \pi_k \left(1 + (\gamma_\varepsilon/\gamma_{s})\dd_k^H\dd_k\right)^{-\frac{1}{2}}\times\\
	      &&\exp\left\lbrace\frac{\gamma_\varepsilon}{2}\operatorname{Re}(\dd_k^H\rr_{i,-k})^2/(\gamma_{s}/\gamma_\varepsilon+\dd_k^H\dd_k) \right\rbrace,\nonumber\\
q_{ik} &\propto & 1 - \pi_k.
\end{eqnarray}
The symbol $H$ denotes the conjugate transpose. The probabilities can be obtained by dividing both of these terms by their sum. We observe that the probability that $z_{ik} = 1$ takes into account how well dictionary element $\dd_k$ correlates with the residual $r_{i,-k}$. After sampling $z_{ik}$ we sample the corresponding weight $s_{ik}$ from its conditional posterior Gaussian distribution,
\begin{equation}
 s_{ik}|z_{ik} \sim N\left(\zz_{ik}\frac{\operatorname{Re}(\dd_k^H\rr_{i,-k})}{\gamma_{s}/\gamma_\varepsilon + \dd_k^H\dd_k},\frac{1}{\gamma_{s} + \gamma_\varepsilon z_{ik}\dd_k^H\dd_k}\right).
\end{equation}
When $z_{ik}=1$, the mean of $s_{ik}$ is the regularized least squares solution and the variance will be small if $\gamma_\varepsilon$ is large. When $z_{ik}=0$, $s_{ik}$ can is sampled from the prior, but does not factor in the model in this case.

\paragraph{Sample $\gamma_\varepsilon$ and $\gamma_{s}$}
We next sample from the conditional gamma posterior distribution of the noise precision and weight precision,
\begin{eqnarray}
 \gamma_\varepsilon \hspace{-8pt}&\sim&\hspace{-8pt} Gam\left(g_0 + \textstyle\frac{1}{2}PN,h_0 + \textstyle\frac{1}{2}\sum_{i}\|R_i\x - D\alpha_i\|_2^2\right),\quad\quad\\
 \gamma_{s} \hspace{-8pt}&\sim&\hspace{-8pt} Gam(e_0 + \textstyle\frac{1}{2}\sum_{i,k} z_{ik}, f_0 + \textstyle\frac{1}{2}\sum_{i,k} z_{ik} s_{ik}^2).
\end{eqnarray}
The expected value of each variable is the first term of the distribution divided by the second, which is close to the inverse of the average empirical error for $\gamma_\varepsilon$.

\paragraph{Sample $\pi_k$}
Sample each $\pi_k$ from its conditional beta posterior distribution,
\begin{equation}
 \pi_k \sim Beta\left(a_0 + \textstyle\sum_{i=1}^N z_{ik}, b_0 + \textstyle\sum_{i=1}^N (1-z_{ik})\right).
\end{equation}
The parameters to the beta distribution include counts of how many times dictionary element $\dd_k$ was used by a patch.\newline

\subsubsection{Algorithm for P3 (MRI reconstruction)}
The final sub-problem is to reconstruct the image $\x$. Our approach takes advantage of the Fourier domain similar to other methods, e.g.\ \cite{ref19,ref44}. The corresponding objective function is
$$\x' = \arg\min_x ~ \sum_{i=1}^N \frac{\rho}{2}\|\psi_i\x - \bbeta_i  + \uu_i\|_2^2 + \sum_{i=1}^N \frac{\gamma_\varepsilon}{2}\|R_i\x - D\alpha_i\|_2^2$$\vspace{-8pt}
$$ + \frac{\lambda}{2}\|\mathcal{F}_u \x-\y\|_2^2.$$
Since this is a least squares problem, $\x$ has a closed form solution that satisfies
\begin{equation}\label{eqn.P3_1}
\left(\rho\Psi^T\Psi + \gamma_\varepsilon \textstyle\sum_i R_i^TR_i + \lambda\mathcal{F}_u^H\mathcal{F}_u\right)\x = ~~~~~~~~~~~~~~~~~~~~~
\end{equation}
$$~~~~~~~~~~~~~~~~~~~~~\rho\Psi^T(\beta-\uu) + \gamma_\varepsilon P \x_{\mbox{\tiny{BPFA}}} + \lambda\mathcal{F}_u^H\y .$$

We recall that $\Psi$ is the matrix of stacked $\psi_i$. The vector $\beta$ is also obtained by stacking each $\bbeta_i$ and $\uu$ is the vector formed by stacking $\uu_i$. The vector $\x_{\mbox{\tiny{BPFA}}}$ is the denoised reconstruction from BPFA using the current $D$ and $\alpha_{1:N}$, which results from the definition $\x_{\mbox{\tiny{BPFA}}} = \frac{1}{P}\textstyle\sum_i R_i^TD\alpha_i$.

We observe that inverting the left $N\times N$ matrix is computationally prohibitive since $N$ is the number of pixels in the image. Fortunately, given the form of the matrix in Equation (\ref{eqn.P3_1}) we can use the procedure described in \cite{ref19} and simplify the problem by working in the Fourier domain. This allows for element-wise updates in $k$-space, followed by an inverse Fourier transform. We represent $\x$ as $\x = \mathcal{F}^H\theta$, where $\theta$ is the Fourier transform of $\x$. We then take the Fourier transform of each side of Equation (\ref{eqn.P3_1}) to give
\begin{equation}
\mathcal{F}\left(\rho\Psi^T\Psi + \gamma_\varepsilon \textstyle\sum_i R_i^TR_i + \lambda\mathcal{F}_u^H\mathcal{F}_u\right)\mathcal{F}^H\theta = ~~~~~~~~~~~~~~
\end{equation}
$$~~~~~~~~~~~~~~~~~~~~~\rho\mathcal{F}\Psi^T(\beta-\uu) + \gamma_\varepsilon\mathcal{F}P\x_{\mbox{\tiny{BPFA}}} + \lambda\mathcal{F}\mathcal{F}_u^H\y .$$
The left-hand matrix simplifies to a diagonal matrix,
\begin{equation}
 \mathcal{F}\left(\rho\Psi^T\Psi + \gamma_\varepsilon \textstyle\sum_i R_i^TR_i + \lambda\mathcal{F}_u^H\mathcal{F}_u\right)\mathcal{F}^H =~~~~~~~~~~~~ 
\end{equation}
$$~~~~~~~~~~~~~~~~~~~~~\rho\Lambda + \gamma_\varepsilon P I_N + \lambda I^u_N.$$
Term-by-term this results as follows: The product of the finite difference operator matrix $\Psi$ with itself yields a circulant matrix, which has the rows of the Fourier matrix $\mathcal{F}$ as its eigenvectors and eigenvalues equal to $\Lambda = \mathcal{F}\Psi^T\Psi\mathcal{F}^H$. The matrix $R_i^TR_i$ is a matrix of all zeros, except for ones on the diagonal entries that correspond to the indices of $\x$ associated with the $i^{th}$ patch. Since each pixel appears in $P$ patches, the sum over $i$ gives $P I_N$, and the Fourier product cancels. The final diagonal matrix $I_N^u$ also contains all zeros, except for ones along the diagonal corresponding to the indices in $k$-space that are measured, which results from $\mathcal{F}\mathcal{F}_u^H\mathcal{F}_u\mathcal{F}^H$.

Since the left matrix is diagonal we can perform element-wise updating of the Fourier coefficients $\theta$,
\begin{equation}\label{eqn.k_space}
 \theta_i = \frac{\rho\mathcal{F}_i\Psi^T(\beta-\uu) + \gamma_\varepsilon P\mathcal{F}_i\x_{\mbox{\tiny{BPFA}}} + \lambda\mathcal{F}_i\mathcal{F}_u^H\y }{\rho\Lambda_{ii} + \gamma_\varepsilon P + \lambda \mathcal{F}_i\mathcal{F}_u^H\boldsymbol{1}}.
\end{equation}
We observe that the rightmost term in the numerator and denominator equals zero if $i$ is not a measured $k$-space location. We invert $\theta$ via the inverse Fourier transform $\mathcal{F}^H$ to obtain the reconstructed MR image $\x'$.

\subsection{Discussion on $\lambda$}\label{sec.lambda}
In noise-free compressed sensing, the fidelity term $\lambda$ can tend to infinity giving an equality constraint for the measured $k$-space values \cite{ref1}. However, when $\y$ is noisy the setting of $\lambda$ is critical for most CS-MRI algorithms since this parameter controls the level of denoising in the reconstructed image.   We note that a feature of dictionary learning CS-MRI approaches is that $\lambda$ can still be set to a very large value, and so parameter selection isn't necessary here. This is because a denoised version of the image is obtained through dictionary learning ($\x_{\mbox{\tiny{BPFA}}}$ in this paper) and can be taken as the denoised reconstruction. In Equation (\ref{eqn.k_space}), we observe that by setting $\lambda$ to a large value, we are effectively fixing the measured $k$-space values and using the $k$-space projection of BPFA and TV to fill in the missing values. The reconstruction $\x$ will be noisy, but have artifacts due to sub-sampling removed. The output image 
$\x_{\mbox{\tiny{BPFA}}}$ is a denoised version of $\x$ using BPFA in essentially the same manner as in Section \ref{sec.denoising}. Therefore, the quality of our algorithm depends largely on the quality of BPFA as an image denoising algorithm \cite{ref34}.  We show examples of this using synthetic and clinical data in Sections \ref{sec.simulated} and \ref{sec.noisyMRI}.

\begin{figure}\centering
\subfigure[Random 25\%]{\includegraphics[width=.155\textwidth]{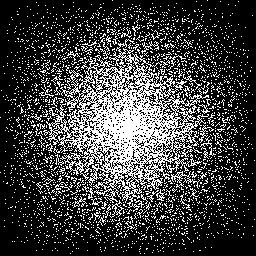}}
\subfigure[Cartesian 30\%]{\includegraphics[width=.155\textwidth]{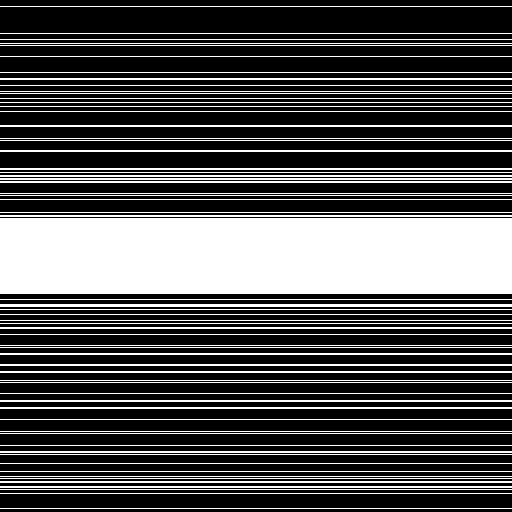}}
\subfigure[Radial 25\%]{\includegraphics[width=.155\textwidth]{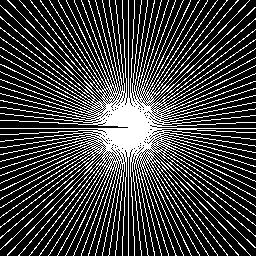}}
\caption{The three masks considered for a given sampling percentage.}\label{fig.mask}
\end{figure}

\section{Experimental Results}\label{sec.experiments}
We evaluate the proposed algorithm on real-valued and complex-valued MRI, and on a synthetic phantom. We consider three sampling masks: 2D random sampling, Cartesian sampling with random phase encodes (1D random), and pseudo radial sampling.\footnote{We used codes referenced in \cite{ref3,ref12,ref14} to generate these masks.} We show an example of each mask in Figure \ref{fig.mask}. We consider a variety of sampling rates for each mask. As a performance measure we use PSNR, and also consider SSIM \cite{ref51}. We compare with three other algorithms: Sparse MRI \cite{ref3}\footnote{\url{http://www.eecs.berkeley.edu/~mlustig/Software.html}}, which as discussed above is a combination of wavelets and total variation, DLMRI \cite{ref19}\footnote{\url{http://www.ifp.illinois.edu/~yoram/DLMRI-Lab/Documentation.
html}}, which is a dictionary learning method based on K-SVD, 
and PBDW
\cite{ref20}\footnote{\url{http://www.quxiaobo.org/index_publications.html}}, which is patch-based method that uses directional wavelets and therefore places greater restrictions on the dictionary. We use the publicly available code for these algorithms indicated above and used the built-in parameter settings, or those indicated in the relevant papers. We also compare with the BPFA algorithm without using total variation by setting $\lambda_g = 0$.

\begin{figure}\centering
\subfigure[Zero filling]{\includegraphics[width=.22\textwidth]{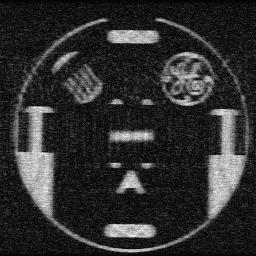}}
\subfigure[BPFA reconstruction ($\x$)]{\includegraphics[width=.22\textwidth]{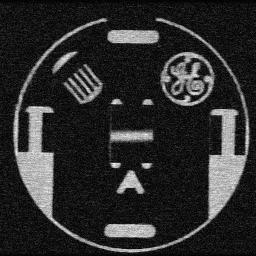}}
\subfigure[BPFA denoising ($\x_{\mbox{\tiny{BPFA}}}$)]{\includegraphics[width=.22\textwidth]{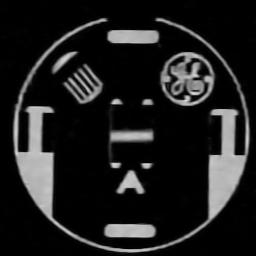}}
\subfigure[Total variation reconstruction]{\includegraphics[width=.22\textwidth]{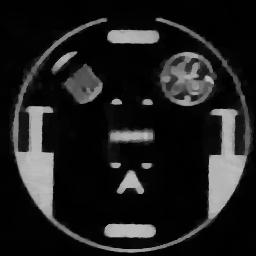}}
\subfigure[Dictionary (magnitude)]{\includegraphics[width=.45\textwidth]{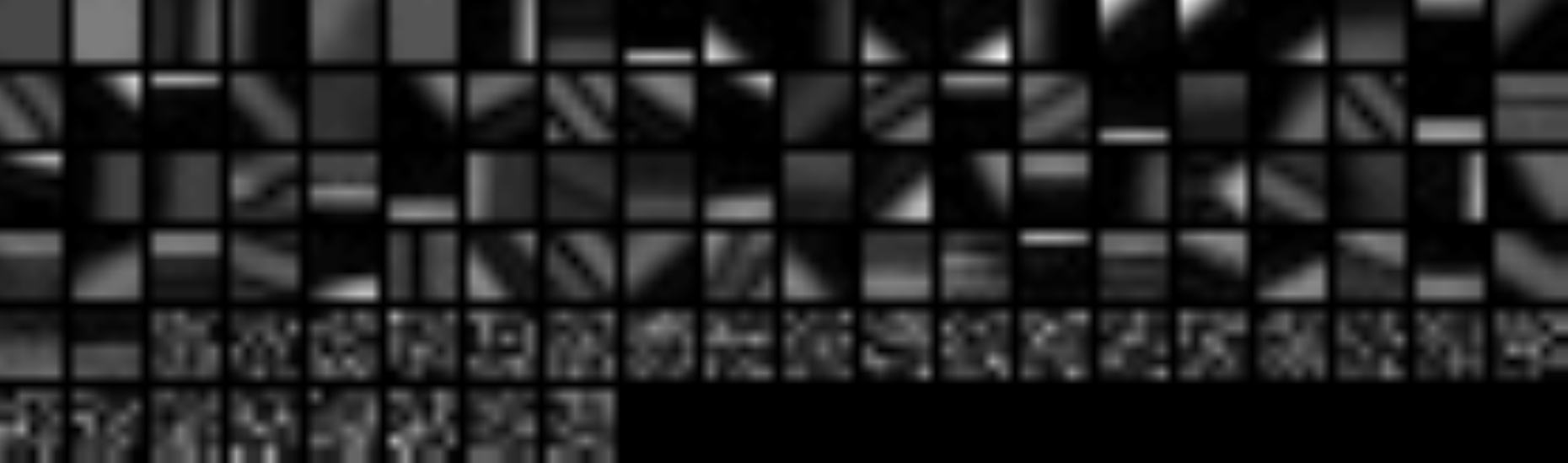}}
\subfigure[Dictionary probabilities]{\includegraphics[width=.22\textwidth]{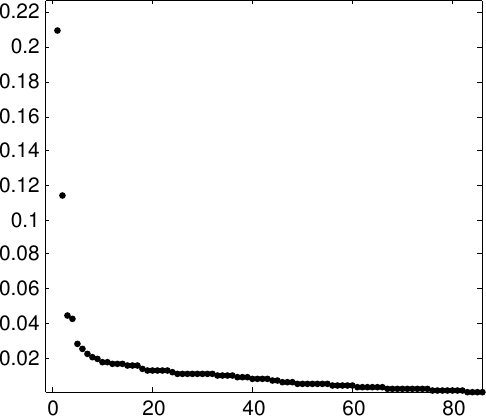}}
\subfigure[Dictionary elements per patch]{\includegraphics[width=.22\textwidth]{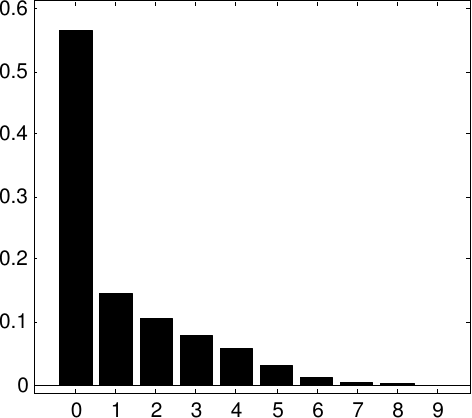}}
    \caption{GE data with noise ($\sigma=0.1$) and 30\% Cartesian sampling. BPFA (b) reconstructs the original noisy image, and (c) denoises the reconstruction simultaneously. (d) TV denoises as part of the reconstruction. Also shown are the dictionary learning variables sorted by $\pi_k$. (e) the dictionary, (f) the distribution on the dictionary, $\pi_k$. (g) The normalized histogram of number of the dictionary elements used per patch.}\label{fig.GEphantom}
\end{figure}

\subsection{Set-up}
For all images, we extract $6\times 6$ patches where each pixel defines the upper left corner of a patch and wrap around the image at the boundaries; we investigate different patch sizes later to show that this is a reasonable size. We initialize $\x$ by zero-filling in $k$-space. We use a dictionary with $K=108$ initial dictionary elements, recalling that the final number of dictionary elements will be smaller due to the sparse BPFA prior. If $108$ is found to be too small, $K$ can be increased with the result being a slower inference algorithm.\footnote{As discussed in Section \ref{sec.BPFA}, in theory $K$ can be infinitely large.} We ran $1000$ iterations and use the results of the last iteration. 

For regularization parameters, we set the data fidelity term $\lambda = 10^{100}$. We are therefore effectively requiring equality with the measured values of $k$-space and allowing BPFA to fill in the missing values, as well as give a denoised reconstruction, as discussed in Section \ref{sec.lambda} and highlighted below in Sections \ref{sec.simulated} and \ref{sec.noisyMRI}. After trying several values, we also found $\lambda_g = 10$ and $\rho = 1000$ to give good results. We set the BPFA hyperparameters as $c = \gamma = e_0=f_0=g_0=h_0=1$. These settings result in a relatively non-informative prior given the amount of data we have. However, we note that our algorithm was robust to these values, since the data overwhelms these prior values when calculating posterior distributions.

\subsection{Experiments on a GE phantom}\label{sec.simulated}
We consider a noisy synthetic example to highlight the advantage of dictionary learning for CS-MRI. In Figure \ref{fig.GEphantom} we show results on a $256\times 256$ GE phantom with additive noise having standard deviation $\sigma = 0.1$. In this experiment we use BPFA without TV to reconstruct the original image using $30\%$ Cartesian sampling. We show the reconstruction using zero-filling in Figure \ref{fig.GEphantom}(a). Since $\lambda = 10^{100}$, we see in Figure \ref{fig.GEphantom}(b) that BPFA essentially helps reconstruct the underlying noisy image for $\x$. However, using the denoising property of the BPFA model shown in Figure \ref{fig:subfig2}, we obtain the denoised reconstruction of Figure \ref{fig.GEphantom}(c) by focusing on $\x_{\tiny\mbox{BPFA}}$ from Equation (\ref{eqn.P3_1}). This is in contrast with the best result we could obtain with TV in Figure \ref{fig.GEphantom}(d), which places the TV penalty on the reconstructed image. As discussed, for TV the setting of $\lambda$ relative to 
$\lambda_g$ is important. We set $\lambda = 1$ and swept through $\lambda_g \in (0,0.15)$, showing the result with highest PSNR in Figure \ref{fig.GEphantom}(d). Similar to Figure \ref{fig:subfig2} we show statistics from the BPFA model in Figures \ref{fig.GEphantom}(e)-(g). We see that roughly 80 dictionary elements were used (the unused noisy elements in Figure \ref{fig.GEphantom}(e) are draws from the prior). We note that $2.28$ elements were used on average by a patch given that at least one was used, which discounts the black regions.

\subsection{Experiments on real-valued (synthetic) MRI}
For our synthetic MRI experiments, we consider two publicly available real-valued $512\times 512$ MRI\footnote{\url{www3.americanradiology.com/pls/web1/wwimggal.vmg/wwimggal.vmg}} of a shoulder and lumbar. We construct these problems by applying the relevant sampling mask to the projection of real-valued MRI into $k$-space. Though using such real-valued MRI data may not reflect clinical reality, we include this idealized setting to provide a complete set of experiments similar to other papers \cite{ref3,ref19,ref20}. We evaluate the performance of our algorithm using PSNR and compare with Sparse MRI \cite{ref3}, DLMRI \cite{ref19} and PBDW \cite{ref20}. Although the original data is real-valued, we learn complex dictionaries since the reconstructions are complex. We consider our algorithm with and without the total variation penalty, denoted BPFA+TV and BPFA, respectively.

\begin{figure}[t]\centering
    \subfigure[BPFA+TV\quad\quad]{\includegraphics[width=.24\textwidth]{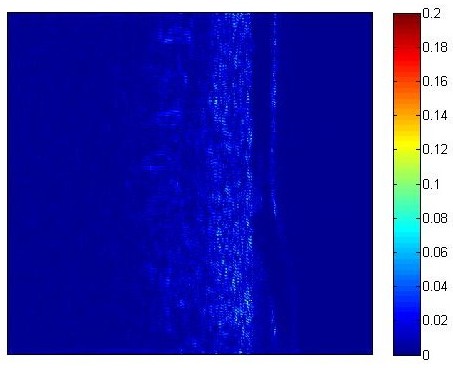}}
    \subfigure[PBDW\quad\quad]{\includegraphics[width=.24\textwidth]{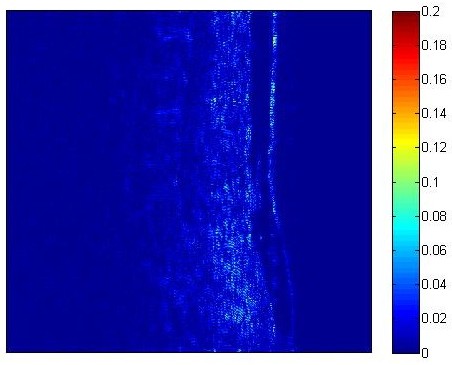}}
    \subfigure[DLMRI\quad\quad]{\includegraphics[width=.24\textwidth]{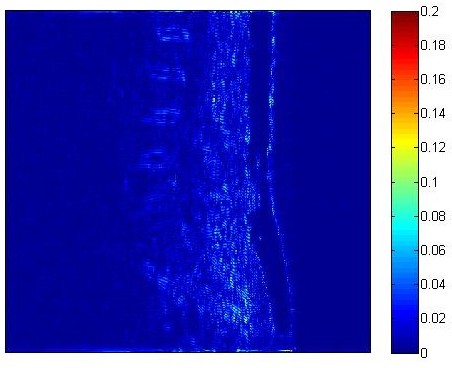}}
    \subfigure[Sparse MRI\quad\quad]{\includegraphics[width=.24\textwidth]{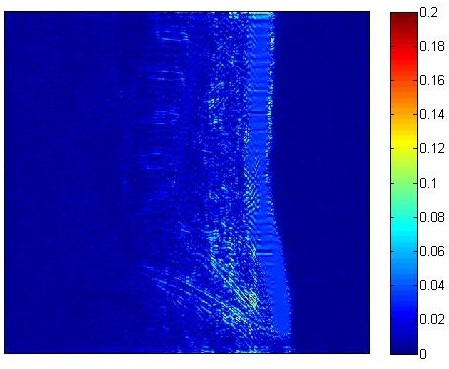}}
    \caption{Absolute errors for 30\% Cartesian sampling of synthetic lumbar MRI.}\label{fig.lumbar_residuals}
\end{figure}

\begin{figure}[t]\centering
    \subfigure[BPFA+TV\quad\quad]{\includegraphics[width=.24\textwidth]{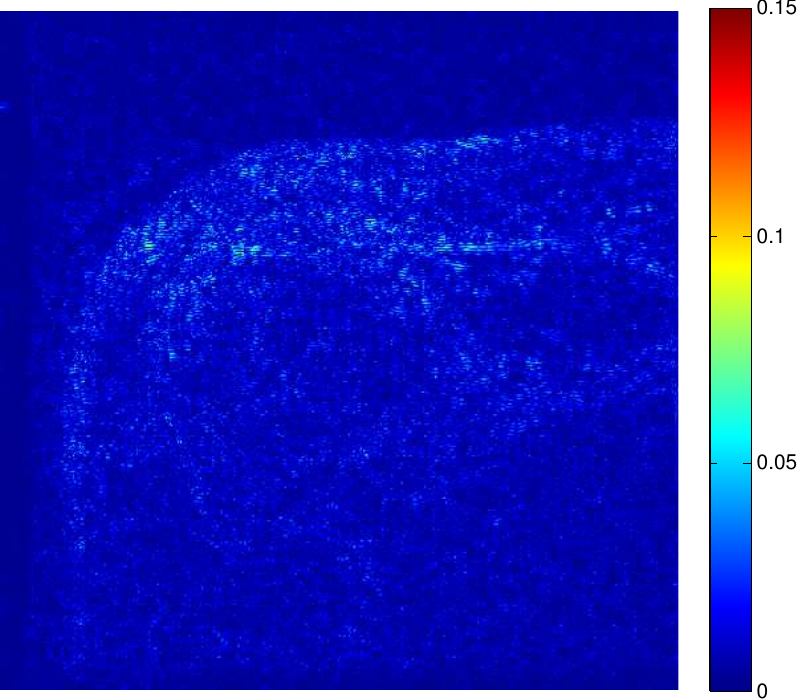}}
    \subfigure[PBDW\quad\quad]{\includegraphics[width=.24\textwidth]{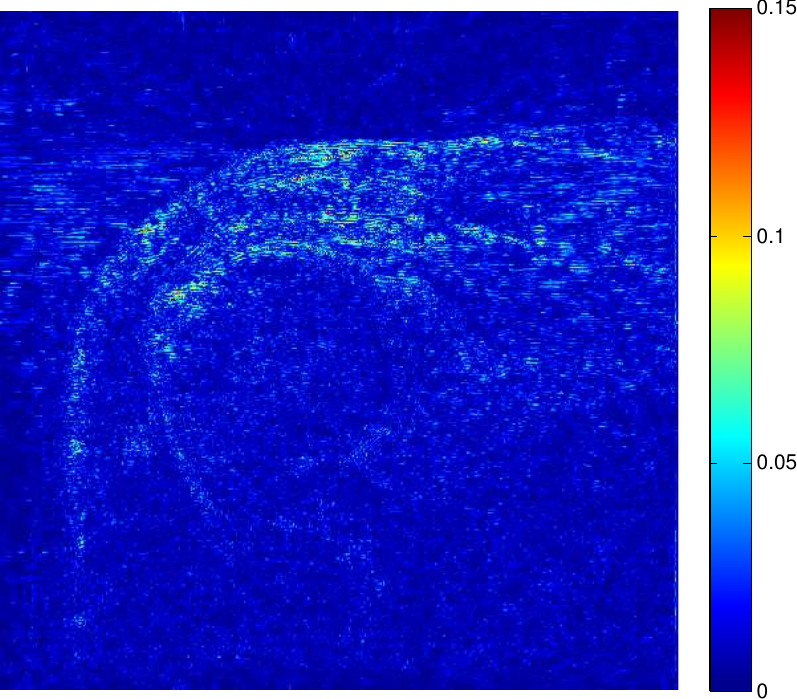}}
    \subfigure[DLMRI\quad\quad]{\includegraphics[width=.24\textwidth]{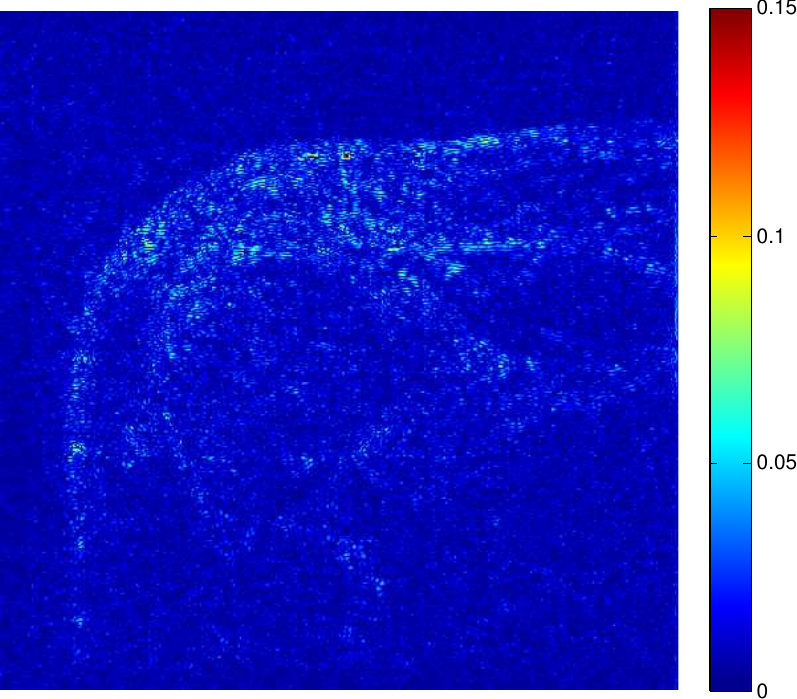}}
    \subfigure[Sparse MRI\quad\quad]{\includegraphics[width=.24\textwidth]{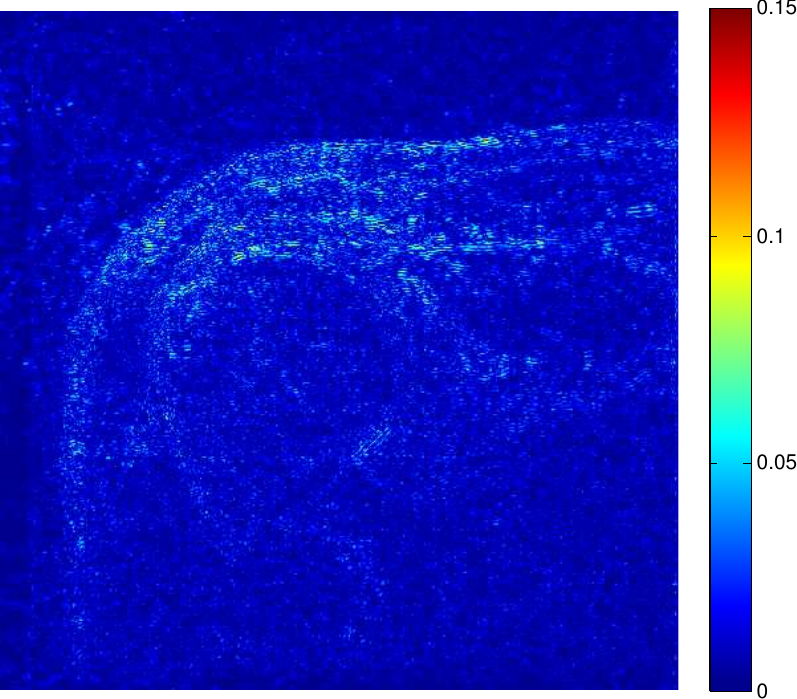}}
    \caption{Absolute errors for 20\% radial sampling of the shoulder MRI.}\label{fig.residuals}
\end{figure}

We present the PSNR results for all sampling masks and rates in Tables \ref{tab.psnr_lumbar} and \ref{tab.psnr_shoulder}. From these values we see the competitive performance of the propose dictionary learning algorithm. We also see a slight improvement by the addition of the TV penalty. As expected, we observe that 2D random sampling produced the best results, followed by pseudo-radial sampling and Cartesian sampling, which is due to their decreasing level of incoherence, with greater incoherence producing artifacts that are more noise-like \cite{ref3}. Since BPFA is good at denoising images, the algorithm naturally performs well in this setting. In Figures \ref{fig.lumbar_residuals} and \ref{fig.residuals} we show the absolute value of the residuals of different algorithms using one experiment from each MRI. We see an improvement using the proposed method, which has more noise-like errors.

\begin{table}
\begin{threeparttable}
\caption{PSNR results for real-valued Lumbar MRI as function of sampling percentage and mask (Cartesian with random phase encodes, 2D random and pseudo radial).}\label{tab.psnr_lumbar}
\begin{tabular}{ rcccccc }
\toprule
\hspace{-7pt} Mask & \hspace{-7pt}Samp\%\hspace{-7pt} & BPFA+TV & BPFA & DLMRI & \hspace{-2pt}SparseMRI\hspace{-2pt} & PBDW\\
\midrule
\hspace{-7pt} {\it Cart.} & \hspace{-7pt}10\hspace{-7pt}  & 32.48  & 32.03  & 31.02  & \hspace{-2pt}30.24\hspace{-2pt}  & 31.74 \\
& \hspace{-7pt}20\hspace{-7pt}  & 36.07  & 35.84  & 33.92  & \hspace{-2pt}33.44\hspace{-2pt}  & 35.19 \\
& \hspace{-7pt}25\hspace{-7pt}  & 38.78  & 38.53  & 36.56  & \hspace{-2pt}35.50\hspace{-2pt}  & 37.43 \\
& \hspace{-7pt}30\hspace{-7pt}  & 41.08  & 40.12  & 38.87  & \hspace{-2pt}35.57\hspace{-2pt}  & 39.23 \\
& \hspace{-7pt}35\hspace{-7pt}  & 41.05  & 40.96  & 38.85  & \hspace{-2pt}37.66\hspace{-2pt}  & ~39.24 \vspace{5pt}\\
\hspace{-7pt} {\it Rand.} & \hspace{-7pt}10\hspace{-7pt}  & 42.82  & 40.81  & 38.25  & \hspace{-2pt}25.87\hspace{-2pt}  & 37.09 \\
& \hspace{-7pt}20\hspace{-7pt}  & 44.35  & 41.80  & 40.11  & \hspace{-2pt}27.80\hspace{-2pt}  & 37.86 \\
& \hspace{-7pt}25\hspace{-7pt}  & 48.11  & 47.09  & 43.51  & \hspace{-2pt}37.22\hspace{-2pt}  & 43.65 \\
& \hspace{-7pt}30\hspace{-7pt}  & 49.36  & 48.55  & 44.93  & \hspace{-2pt}38.72\hspace{-2pt}  & 45.50 \\
& \hspace{-7pt}35\hspace{-7pt}  & 50.20  & 49.19  & 45.87  & \hspace{-2pt}41.70\hspace{-2pt}  & ~46.85 \vspace{5pt}\\
\hspace{-7pt} {\it Rad.} & \hspace{-7pt}10\hspace{-7pt}  & 35.16  & 33.33  & 32.91  & \hspace{-2pt}29.35\hspace{-2pt}  & 31.46 \\
& \hspace{-7pt}20\hspace{-7pt}  & 41.69  & 41.18  & 38.38  & \hspace{-2pt}35.69\hspace{-2pt}  & 38.01 \\
& \hspace{-7pt}25\hspace{-7pt}  & 43.75  & 43.40  & 40.29  & \hspace{-2pt}38.59\hspace{-2pt}  & 40.25 \\
& \hspace{-7pt}30\hspace{-7pt}  & 45.22  & 44.95  & 41.86  & \hspace{-2pt}37.37\hspace{-2pt}  & 42.11 \\
& \hspace{-7pt}35\hspace{-7pt}  & 46.85  & 46.45  & 43.09  & \hspace{-2pt}39.74\hspace{-2pt}  & 43.72 \\
\bottomrule
\end{tabular}
\end{threeparttable}
\end{table}

\begin{table}
\begin{threeparttable}
\caption{PSNR results for real-valued Shoulder MRI as function of sampling percentage and mask (Cartesian with random phase encodes, 2D random and pseudo radial).}\label{tab.psnr_shoulder}
\begin{tabular}{ rcccccc }
\toprule
\hspace{-7pt}Mask & \hspace{-7pt}Samp\%\hspace{-7pt} & BPFA+TV & BPFA & DLMRI & \hspace{-2pt}SparseMRI\hspace{-2pt} & PBDW\\
\midrule
\hspace{-7pt}{\it Cart.} & \hspace{-7pt}10\hspace{-7pt}  & 32.65  & 30.79  & 31.02  & \hspace{-2pt}27.65\hspace{-2pt}  & 28.88 \\
& \hspace{-7pt}20\hspace{-7pt}  & 36.96  & 35.77  & 34.52  & \hspace{-2pt}30.64\hspace{-2pt}  & 32.10 \\
& \hspace{-7pt}25\hspace{-7pt}  & 38.45  & 37.97  & 35.69  & \hspace{-2pt}32.44\hspace{-2pt}  & 34.12 \\
& \hspace{-7pt}30\hspace{-7pt}  & 41.43  & 41.22  & 38.11  & \hspace{-2pt}34.26\hspace{-2pt}  & 36.73 \\
& \hspace{-7pt}35\hspace{-7pt}  & 41.33  & 41.14  & 38.44  & \hspace{-2pt}34.50\hspace{-2pt}  &~36.76  \vspace{5pt}\\
\hspace{-7pt}{\it Rand.} & \hspace{-7pt}10\hspace{-7pt}  & 41.00  & 39.96  & 38.18  & \hspace{-2pt}30.72\hspace{-2pt}  & 36.48 \\
& \hspace{-7pt}20\hspace{-7pt}  & 43.53  & 42.40  & 39.38  & \hspace{-2pt}32.08\hspace{-2pt}  & 39.39 \\
& \hspace{-7pt}25\hspace{-7pt}  & 45.43  & 45.44  & 42.58  & \hspace{-2pt}40.81\hspace{-2pt}  & 41.31 \\
& \hspace{-7pt}30\hspace{-7pt}  & 46.89  & 46.86  & 44.03  & \hspace{-2pt}43.47\hspace{-2pt}  & 43.12 \\
& \hspace{-7pt}35\hspace{-7pt}  & 47.95  & 47.87  & 45.01  & \hspace{-2pt}44.89\hspace{-2pt}  &~44.45 \vspace{5pt}\\
\hspace{-7pt}{\it Rad.} & \hspace{-7pt}10\hspace{-7pt}  & 34.30  & 33.88  & 33.27  & \hspace{-2pt}29.18\hspace{-2pt}  & 31.60 \\
& \hspace{-7pt}20\hspace{-7pt}  & 39.41  & 39.47  & 38.06  & \hspace{-2pt}35.50\hspace{-2pt}  & 36.38 \\
& \hspace{-7pt}25\hspace{-7pt}  & 41.40  & 41.52  & 39.73  & \hspace{-2pt}38.70\hspace{-2pt}  & 38.21 \\
& \hspace{-7pt}30\hspace{-7pt}  & 43.14  & 43.45  & 41.20  & \hspace{-2pt}39.98\hspace{-2pt}  & 40.30 \\
& \hspace{-7pt}35\hspace{-7pt}  & 44.69  & 44.99  & 42.58  & \hspace{-2pt}39.11\hspace{-2pt}  & 41.72 \\
\bottomrule
\end{tabular}
\end{threeparttable}
\end{table}

\begin{table*}
\centering
\begin{threeparttable}
\caption{PSNR/SSIM results for complex-valued Brain MRI as a function of sampling percentage. Sampling masks include Cartesian sampling with random phase encodes, 2D random sampling and pseudo radial sampling.}\label{tab.psnr_ssim_brain}
\begin{tabular}{ rccccccc }
\toprule
Mask~ & Sample \% & BPFA+TV & BPFA & DLMRI & Sparse MRI & PBDW & Zero-filling\\
\midrule
{\it Cartesian} & 25 & ~~35.62~/~0.951~~ & ~~34.86~/~0.948~~ & ~~29.90~/~0.812~~ & ~~25.29~/~0.696~~ & ~~34.69~/~0.935~~ &  24.13~/~0.591 \\
& 30 & 38.64~/~0.968 & 37.70~/~0.965 & 31.54~/~0.849 & 26.16~/~0.745 & 37.24~/~0.957 & 24.55~/~0.614 \\
& 35 & 39.36~/~0.972 & 38.87~/~0.971 & 32.35~/~0.863 & 27.35~/~0.795 & 37.90~/~0.963 & 24.94~/~0.616 \\
& 40 & 41.09~/~0.977 & 40.45~/~0.976 & 33.60~/~0.876 & 29.82~/~0.845 & 39.23~/~0.969 & ~26.28~/~0.667 \vspace{5pt}\\
{\it Random} & 10 & 31.57~/~0.923  & 31.24~/~0.920  & 29.38~/~0.821  & 24.85~/~0.756  & 31.15~/~0.921 & 23.23~/~0.536 \\
& 15 & 36.49~/~0.963  & 35.44~/~0.961  & 30.16~/~0.774  & 22.68~/~0.651  & 34.22~/~0.942 & 21.18~/~0.493 \\
& 20 & 38.83~/~0.962  & 38.38~/~0.964  & 31.62~/~0.804  & 26.28~/~0.672  & 36.29~/~0.960 & 23.52~/~0.504 \\
& 25 & 40.75~/~0.979  & 40.00~/~0.973  & 32.83~/~0.862  & 31.16~/~0.934  & 37.62~/~0.968 & 26.58~/~0.582 \\
& 30 & 42.70~/~0.984  & 42.24~/~0.984  & 34.09~/~0.887  & 31.90~/~0.965  & 39.38~/~0.976 & ~27.67~/~0.630 \vspace{5pt}\\
{\it Radial} & 10 & 30.76~/~0.914  & 30.68~/~0.914  & 27.78~/~0.680  & 19.79~/~0.482 & 30.78~/~0.886 & 19.06~/~0.367  \\
& 15 & 34.00~/~0.949  & 33.79~/~0.950  & 29.49~/~0.734  & 22.07~/~0.640 & 33.99~/~0.937 & 20.87~/~0.498  \\
& 20 & 36.92~/~0.967  & 36.60~/~0.967  & 30.78~/~0.768  & 24.22~/~0.739 & 36.34~/~0.958 & 22.57~/~0.537  \\
& 25 & 39.72~/~0.977  & 39.37~/~0.977  & 31.91~/~0.794  & 26.64~/~0.797 & 38.38~/~0.970 & 24.34~/~0.574  \\
& 30 & 41.81~/~0.982  & 41.54~/~0.982  & 32.77~/~0.807  & 28.20~/~0.827 & 39.74~/~0.975 & 25.43~/~0.600  \\
\bottomrule
\end{tabular}
\end{threeparttable}
\end{table*}

\begin{figure*}\centering
    \subfigure[Original]{\includegraphics[width=.192\textwidth]{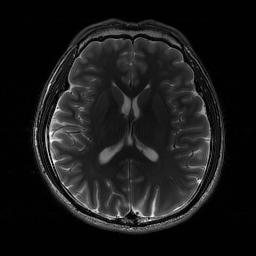}}
    \subfigure[BPFA+TV]{\includegraphics[width=.192\textwidth]{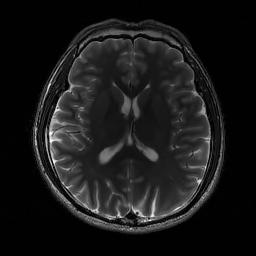}}
    \subfigure[BPFA]{\includegraphics[width=.192\textwidth]{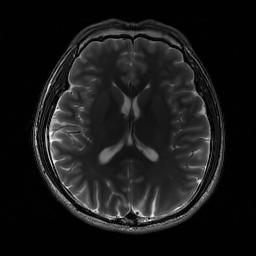}}
    \subfigure[PBDW]{\includegraphics[width=.192\textwidth]{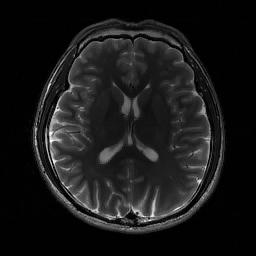}}
    \subfigure[DLMRI]{\includegraphics[width=.192\textwidth]{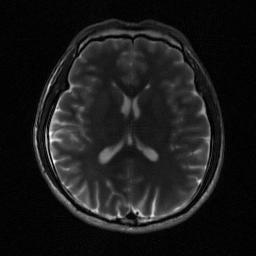}}
    \subfigure[PSNR vs iteration]{\includegraphics[width=.192\textwidth]{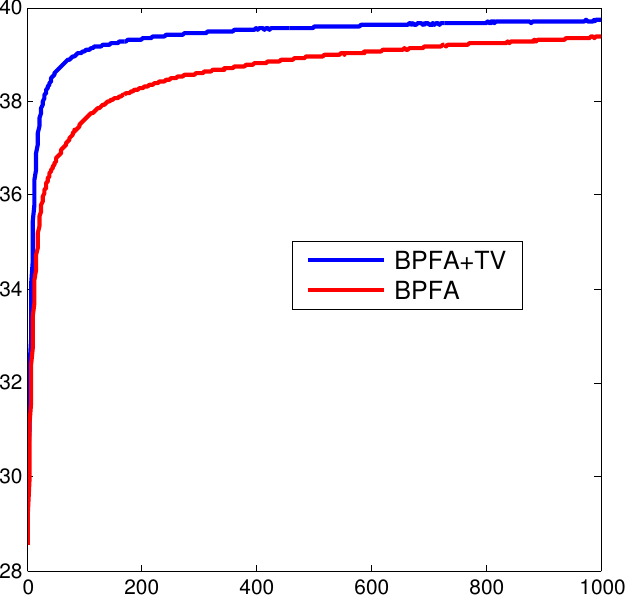}}
    \subfigure[BPFA+TV error\quad\quad]{\includegraphics[width=.192\textwidth]{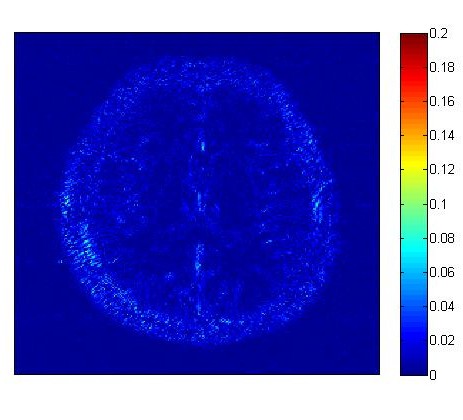}}
    \subfigure[BPFA error\quad\quad]{\includegraphics[width=.192\textwidth]{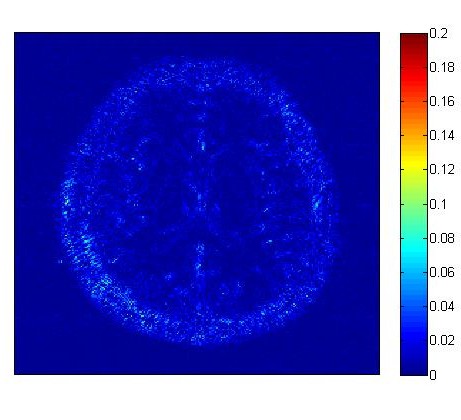}}
    \subfigure[PBDW error\quad\quad]{\includegraphics[width=.192\textwidth]{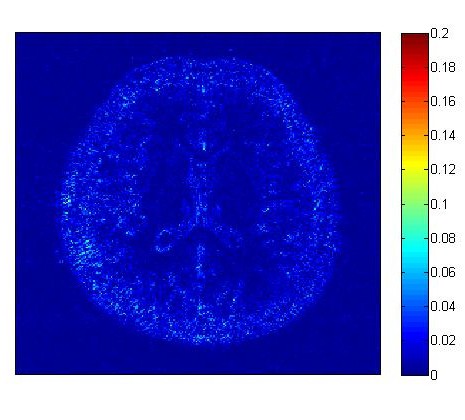}}
    \subfigure[DLMRI error\quad\quad]{\includegraphics[width=.192\textwidth]{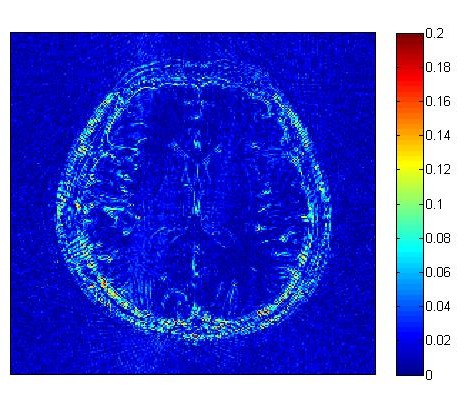}}
    \caption{Reconstruction results for 25\% pseudo radial sampling of a complex-valued MRI of the brain.}\label{fig.complex_brain}
\end{figure*}

\begin{figure*}\centering
    \subfigure[Original]{\includegraphics[width=.192\textwidth]{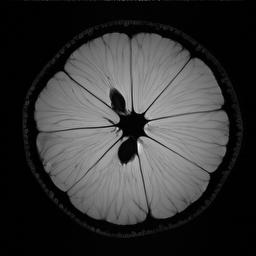}}
    \subfigure[BPFA+TV: PSNR = 39.64]{\includegraphics[width=.192\textwidth]{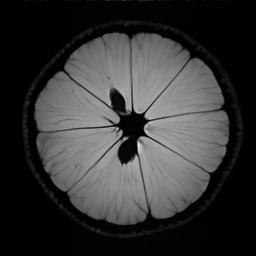}}
    \subfigure[BPFA: PSNR = 38.21]{\includegraphics[width=.192\textwidth]{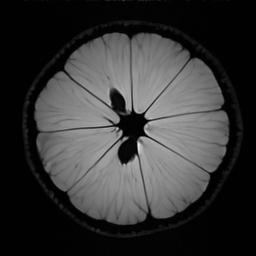}}
    \subfigure[PBDW: PSNR = 37.89]{\includegraphics[width=.192\textwidth]{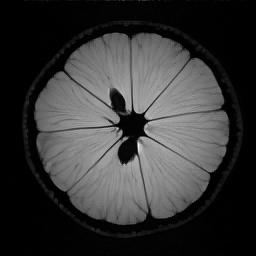}}
    \subfigure[DLMRI: PSNR = 35.05]{\includegraphics[width=.192\textwidth]{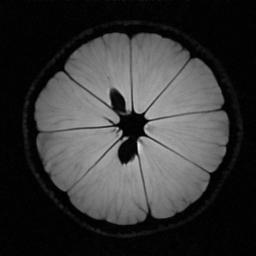}}
    \subfigure[PSNR vs iteration]{\includegraphics[width=.192\textwidth]{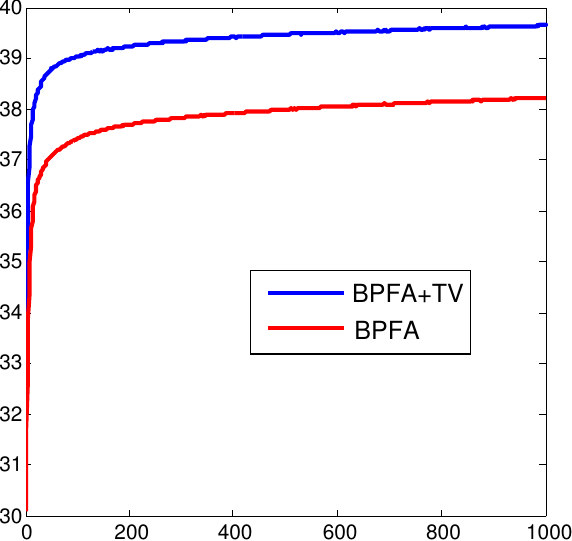}}
    \subfigure[BPFA+TV error\quad\quad]{\includegraphics[width=.192\textwidth]{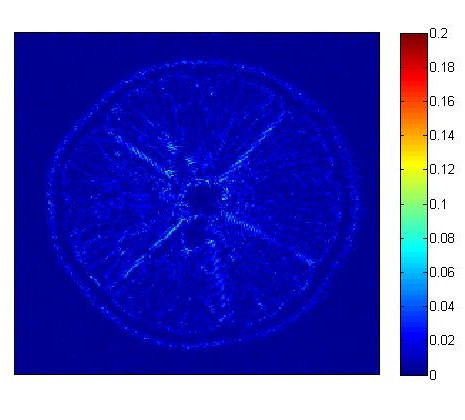}}
    \subfigure[BPFA error\quad\quad]{\includegraphics[width=.192\textwidth]{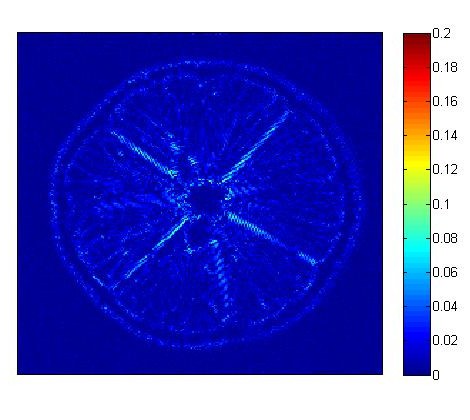}}
    \subfigure[PBDW error\quad\quad]{\includegraphics[width=.192\textwidth]{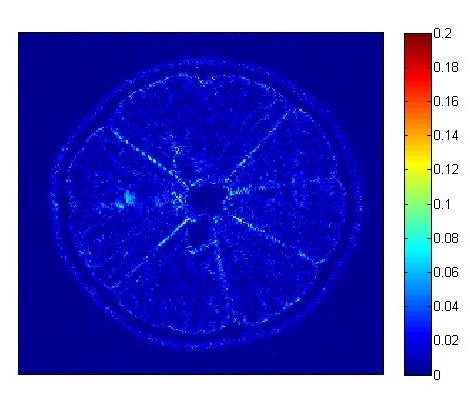}}
    \subfigure[DLMRI error\quad\quad]{\includegraphics[width=.192\textwidth]{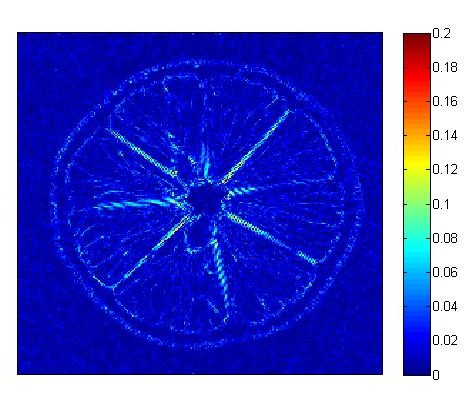}}
    \caption{Reconstruction results for 35\% 2D random sampling of a complex-valued MRI of a lemon.}\label{fig.complex_lemon}
\end{figure*}

\subsection{Experiments on complex-valued MRI}
We also consider two clinically obtained complex-valued MRI: We use the T2-weighted brain MRI from \cite{ref6}, which is a $256 \times 256$ MRI of a healthy volunteer from a 3T Siemens Trio Tim MRI scanner using the T2-weighted turbo spin echo sequence (TR/TE = 6100/99 ms, 220 $\times$ 220 mm field of view, 3 mm slice thickness).  We also use an MRI scan of a lemon obtained from the Research Center of Magnetic Resonance and Medical Imaging at Xiamen University (TE = 32 ms, size = 256 $\times$ 256, spin echo sequence, TR/TE=10000/32 ms, FOV= 70$\times$70 mm$^2$, 2-mm slice thickness). This MRI is from a 7T/160mm bore Varian MRI system (Agilent Technologies, Santa Clara, CA, USA) using a quadrature-coil probe.

For the brain MRI experiment we use both PSNR and SSIM as performance measures. We show these values in Table \ref{tab.psnr_ssim_brain} for each algorithm, sampling mask and sampling rate. As with the synthetic MRI, we see that our algorithm performs competitively with the state-of-the-art. We also see the significant improvement of all algorithms over zero-filling. Example reconstructions are shown for each MRI dataset in Figures \ref{fig.complex_brain} and \ref{fig.complex_lemon}. Also in Figure \ref{fig.complex_lemon} are PSNR values for the lemon MRI. We see from the absolute error residuals for these experiments that the BPFA algorithm learns a slightly finer detail structure compared with other algorithms, with the errors being more noise-like. We also show the PSNR of BPFA+TV and BPFA as a function of iteration. As is evident, the algorithm does not necessarily need all 1000 iterations, but performs competitively even in half that number.

\begin{figure}\centering
\includegraphics[width=.45\textwidth]{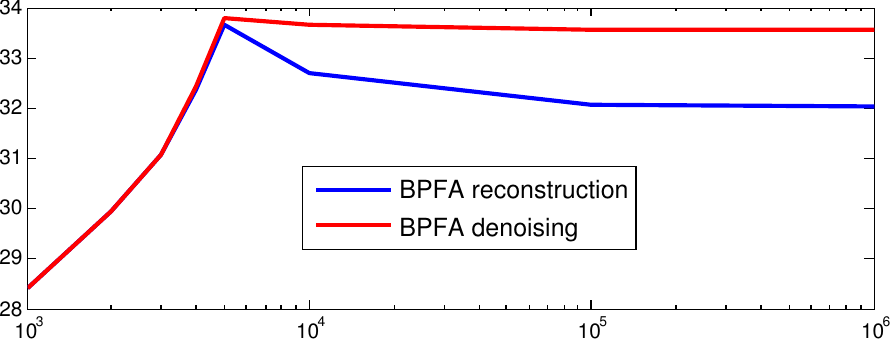}
\caption{PSNR vs $\lambda$ in the noisy setting ($\sigma = 0.03$) for the complex-value brain MRI with 30\% 2D random sampling.}\label{fig.psnr_vs_lambda_noisy}
\end{figure}

\begin{figure}\centering
\subfigure[Zero filling]{\includegraphics[width=.22\textwidth]{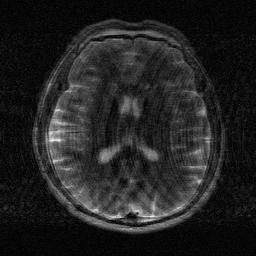}}
\subfigure[BPFA reconstruction ($\x$)]{\includegraphics[width=.22\textwidth]{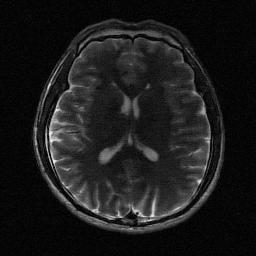}}
\subfigure[BPFA denoising ($\x_{\mbox{\tiny{BPFA}}}$)]{\includegraphics[width=.22\textwidth]{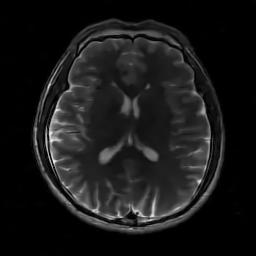}}
\subfigure[DLMRI]{\includegraphics[width=.22\textwidth]{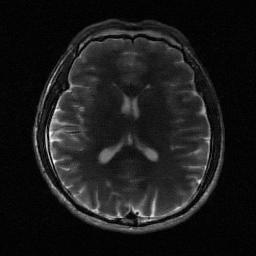}}
\caption{The denoising properties of dictionary learning on noisy complex-valued MRI with 35\% Cartesian sampling and $\sigma = 0.03$.}\label{fig.complex_brain_noisy}
\end{figure}

\subsection{Experiments in the noisy setting}\label{sec.noisyMRI}

The MRI we have considered thus far have been essentially noiseless. For some MRI machines this may be an unrealistic assumption. We continue our evaluation of noisy MRI begun with the toy GE phantom in Section \ref{sec.simulated} by evaluating how our model performs on clinically obtained MRI with additive noise. We show BPFA results without TV to highlight the dictionary learning features, but note that results with TV provide a slight improvement in terms of PSNR and SSIM. We again consider the brain MRI and use additive complex white Gaussian noise having standard deviation $\sigma = 0.01,0.02,0.03$. For all experiments we use the original noise-free MRI as the ground truth. 

As discussed in Section \ref{sec.lambda} and illustrated in Section \ref{sec.simulated}, dictionary learning allows us to consider two possible reconstructions: the actual reconstruction $\x$, and the denoised BPFA reconstruction $\x_{\mbox{\tiny{BPFA}}} = \frac{1}{P}\sum_i R_i^T D\alpha_i$. As detailed in these sections, as $\lambda$ becomes larger the reconstruction will be noisier, but with the artifacts from sub-sampling removed. However, for all values of $\lambda$, $\x_{\mbox{\tiny{BPFA}}}$ produces a denoised version that essentially doesn't change. We see this clearly in Figure \ref{fig.psnr_vs_lambda_noisy}, where we show the PSNR of each reconstruction as a function of $\lambda$. When $\lambda$ is small, the performance degrades for both algorithms since too much smoothing is done by dictionary learning on $\x$. As $\lambda$ increases, both improve, but eventually the reconstruction of $\x$ degrades again because near equality to the noisy $\y$ is being more strictly enforced. The denoised 
reconstruction 
however levels off and does not degrade. We show PSNR values in Table \ref{tab.psnr_noise} as a function of noise level.\footnote{We are working with a different scaling of the MRI than in \cite{ref19} and made the appropriate modifications. Also, since DLMRI is a dictionary learning method it can output ``$\x_{\mbox{\tiny{KSVD}}}$'', though it was not originally motivated this way. Issues discussed in Sections \ref{sec.KSVD} and \ref{sec.denoising} apply in this case.} Example reconstructions that parallel those given in Figure \ref{fig.GEphantom} are also shown in Figure \ref{fig.complex_brain_noisy}. These results highlight the robustness of our approach to $\lambda$ in the noisy setting, and we note that we encountered no stability issues using extremely large values of $\lambda$.

\begin{table}
\centering
\begin{threeparttable}
\caption{PSNR for 35\% Cartesian sampling of complex-valued Brain MRI for various noise standard deviations. ($\lambda = 10^{100}$)}\label{tab.psnr_noise}
\begin{tabular}{ ccccc }
\toprule
Reconstruction method & $\sigma = 0$ & $\sigma = 0.01$ & $\sigma = 0.02$ & $\sigma = 0.03$\\
\midrule
BPFA--reconstruction & 38.87 & 37.25 & 33.77 & 31.08 \\
BPFA--denoising & 37.99 & 37.19 & 34.43 & 32.39 \\
DLMRI  & 32.35 & 32.12 & 31.61 & 30.65 \\
\bottomrule
\end{tabular}
\end{threeparttable}
\end{table}

\subsection{Dictionary learning and further discussion}

\begin{figure}[t]\centering
    \subfigure[Dictionary (magnitude) for 10\% sampling]{\includegraphics[width=.48\textwidth]{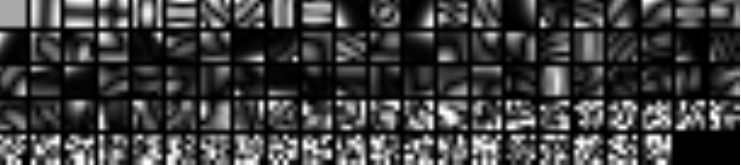}}
    \subfigure[Dictionary (magnitude) for 20\% sampling]{\includegraphics[width=.48\textwidth]{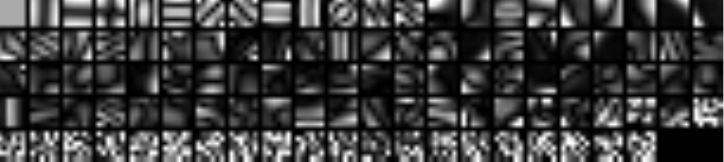}}
    \subfigure[Dictionary (magnitude) for 30\% sampling]{\includegraphics[width=.48\textwidth]{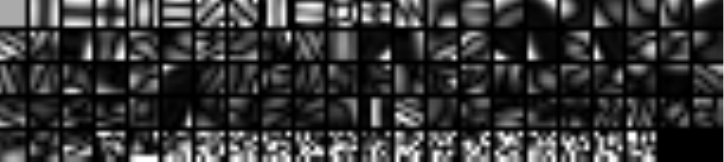}}
    \subfigure[BPFA weights (cumulative)]{\includegraphics[height=1.34in]{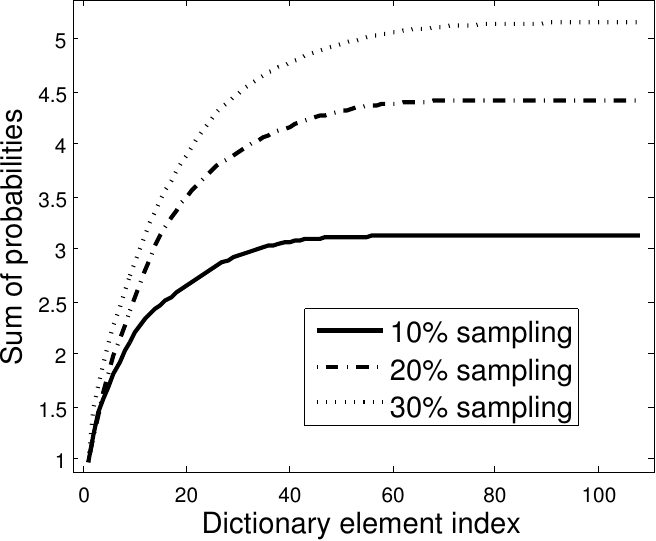}}\,
    \subfigure[Dictionary elements per patch]{\includegraphics[height=1.34in]{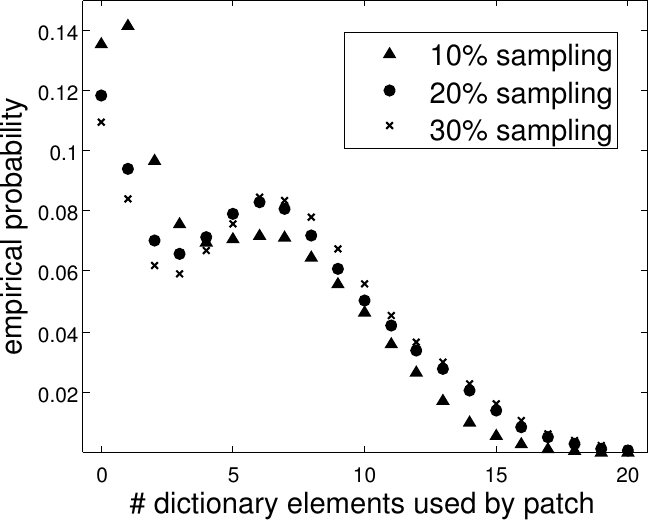}}
    \caption{Radial sampling for the Brain MRI. (a)-(c) The learned dictionary for various sampling rates. The noisy elements towards the end of each were unused and are samples from the prior. (d) The cumulative function of the sorted $\pi_k$ from BPFA for each sampling rate. This gives information on sparsity and average usage of the dictionary. (e) The distribution on the number of elements used per patch for each sampling rate.}\label{fig.example_bpfa}
\end{figure}

We investigate the model learned by BPFA. In Figure \ref{fig.example_bpfa} we show dictionary learning results learned by BPFA+TV for radial sampling of the complex Brain MRI. In the top portion, we show the dictionaries learned for 10\%, 20\% and 30\% sampling. We see that they are similar in their shape, but the number of elements increases as the sampling percentage increases since more complex information about the image is contained in the $k$-space measurements. We again note that unused elements are represented by draws from the prior. In Figure \ref{fig.example_bpfa}(d) we show the cumulative sum of the ordered $\pi_k$ from BPFA. We can read off the average number of elements used per patch by looking at the right-most value. We see that more elements are used per patch as the fraction of observed $k$-space increases. We also see that for 10\%, 20\% and 30\% sampling, roughly 70, 80 and 95, respectively, of the 108 total dictionary elements were significantly 
used, as indicated by the leveling off of these functions. This highlights the adaptive property of the nonparametric beta process prior. In Figure \ref{fig.example_bpfa}(e) we show the empirical distribution on the number of dictionary elements used per patch for each sampling rate. We see that there are two modes, one for the empty background and one for the foreground, and the second mode tends to increase as the sampling rate increases. The adaptability of this value to each patch is another characteristic of the beta process model.

\begin{table}
\centering
\begin{threeparttable}
\caption{PSNR as a function of patch size for a real-valued and complex-valued Brain MRI with Cartesian sampling.}\label{tab.patch_size}
\begin{tabular}{ cccccc }
\toprule
 & 4$\times$4 & 5$\times$5 & 6$\times$6 & 7$\times$7 & 8$\times$8\\
\midrule
Synthetic brain 25\% & 37.86 & 38.29 & 38.33 & 38.26 & 38.24 \\
Complex brain 40\% & 40.53 & 40.84 & 41.09 & 41.11 & 41.15 \\
\bottomrule\vspace{5pt}
\end{tabular}
\end{threeparttable}

\begin{threeparttable}
\caption{Total runtime in minutes (seconds/iteration). We ran 1000 iterations of BPFA, 100 of DLMRI and 10 of Sparse MRI.}\label{tab.run_time}
\begin{tabular}{ ccccc }
\toprule
Sampling \% & BPFA+TV & BPFA & DLMRI & Sparse MRI \\
\midrule
10\% & 52.4 (3.15) & 50.5 (3.03) & 27.6 (16.5) & 1.63 (9.78)\\
20\% & 51.3 (3.08) & 49.5 (2.97) & 38.3 (23.0) & 1.59 (9.54)\\
30\% & 51.2 (3.07) & 48.3 (2.90) & 45.7 (27.4) & 1.60 (9.60)\\
\bottomrule
\end{tabular}
\end{threeparttable}
\end{table}

We also performed an experiment with varying patch sizes and show our results in Table \ref{tab.patch_size}. We see that the results are not very sensitive to this setting and that comparisons using $6\times 6$ patches are meaningful.  We also compare the runtime for different algorithms in Table \ref{tab.run_time}, showing both the total runtime of each algorithm and the per-iteration times using an Intel Xeon CPU E5-1620 at 3.60GHz, 16.0G ram. However, we note that we arguably ran more iterations than necessary for these algorithms; the BPFA algorithms generally produced high quality results in half the number of iterations, as did DLMRI (the authors of \cite{ref19} recommend 20 iterations), while Sparse MRI uses 5 iterations as default and the performance didn't improve beyond 10 iterations. We note that the speed-up over DLMRI arises from the lack of the OMP algorithm, 
which in Matlab is much slower than our sparse coding update.\footnote{BPFA is significantly faster than K-SVD in Matlab because it requires fewer loops. This difference may not be as large with other coding languages.} We note that inference for the BPFA model is easily parallelizable---as are the other dictionary learning algorithms---which can speed up processing time.

The proposed method has several advantages, which we believe leads to the improvement in performance. A significant advantage is the adaptive learning of the dictionary size and per-patch sparsity level using a nonparametric stochastic process that is naturally suited for this problem. Several other dictionary learning parameters such as the noise variance and the variances of the score weights are adjusted as well through a natural MCMC sampling approach. These benefits have been investigated in other applications of this model \cite{ref34}, and naturally translate here since CS-MRI with BPFA is closely related to image denoising as we have shown.

Another advantage of our model is the Markov Chain Monte Carlo inference algorithm itself. In highly non-convex Bayesian models (or similar models with a Bayesian interpretation), it is often observed by the statistics community that MCMC sampling can outperform deterministic methods, and rarely performs worse \cite{ref53}. Given that BPFA is a Bayesian model, such sampling techniques are readily derived, as we showed in Section \ref{sec.algorithm}.

\section{Conclusion}
We have presented an algorithm for CS-MRI reconstruction that uses Bayesian nonparametric dictionary learning. Our Bayesian approach uses a model called beta process factor analysis (BPFA) for {\it in situ} dictionary learning. Through this hierarchical generative structure, we can learn the dictionary size, sparsity pattern and additional regularization parameters. We also considered a total variation penalty term for additional constraints on image smoothness. We presented an optimization algorithm using the alternating direction method of multipliers (ADMM) and MCMC Gibbs sampling for all BPFA variables. Experimental results on real and complex-valued MRI showed that our proposed regularization framework compares favorably with other algorithms for various sampling trajectories and rates. We also showed the natural ability of dictionary learning to handle noisy MRI {\it without} dependence on the measurement fidelity parameter $\lambda$. To this end, we showed that the model can enforce a near equality 
constraint to the noisy measurements and use the dictionary learning result as a denoised output of the noisy MRI.

\section{Appendix}
We give a brief review of the ADMM algorithm \cite{ref46}. We start with the convex optimization problem
\begin{equation}\label{eqn.admm_generic}
 \min_x ~\|A\x - b\|_2^2 + h(\x),
\end{equation}
where $h$ is a non-smooth convex function, such as an $\ell_1$ penalty. ADMM decouples the smooth squared error term from this penalty by introducing a second vector $\vv$ such that
\begin{equation}
 \min_x ~\|A\x - b\|_2^2 + h(\vv) ~~ \mbox{subject to}~~ \vv = \x.
\end{equation}
This is followed by a relaxation of the equality $\vv = \x$ via an augmented Lagrangian term
\begin{equation}\label{eqn.aug_lag}
 L(\x,\vv,\eta) = \|A\x - b\|_2^2 + h(\vv) + \eta^T(\x - \vv) + \frac{\rho}{2}\|\x-\vv\|_2^2.
\end{equation}
A minimax saddle point is found with the minimization taking place over both $\x$ and $\vv$ and dual ascent for $\eta$.

Another way to write the objective in (\ref{eqn.aug_lag}) is to define $\uu = (1/\rho)\eta$ and combine the last two terms. The result is an objective that can be optimized by cycling through the following updates for $\x$, $\vv$ and $\uu$,
\begin{eqnarray}
 \x' & = & \arg\min_x ~\|A\x-b\|_2^2 + \frac{\rho}{2}\|\x - \vv + \uu\|_2^2,\\
\vv' & = & \arg\min_v ~h(\vv) + \frac{\rho}{2}\|\x' - \vv + \uu\|_2^2,\\
\uu' & = & \uu + \x' - \vv'.
\end{eqnarray}
This algorithm simplifies the optimization since the objective for $\x$ is quadratic and thus has a simple analytic solution, while the update for $\vv$ is a proximity operator of $h$ with penalty $\rho$, the difference being that $\vv$ is not pre-multiplied by a matrix as $\x$ is in (\ref{eqn.admm_generic}). Such objective functions tend to be easier to optimize. For example when $h$ is the TV penalty the solution for $\vv$ is analytical.

\begin{IEEEbiography}[{\includegraphics[width=1in,height=1.25in,clip,keepaspectratio]{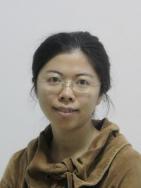}}]{Yue Huang}
 received the B.S. degree in Electrical Engineering from Xiamen University in 2005, and the Ph.D. degree in Biomedical Engineering from Tsinghua University in 2010. Since 2010 she is an Assistant Professor of the School of Information Science and Engineering at Xiamen University. Her main research interests include image processing, machine learning, and biomedical engineering. 
\end{IEEEbiography}

\begin{IEEEbiography}[{\includegraphics[width=1in,height=1.25in,clip,keepaspectratio]{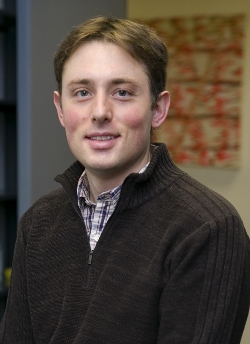}}]{John Paisley}
is an assistant professor in the Department of Electrical Engineering at Columbia University. Prior to that he was a postdoctoral researcher in the Computer Science departments at UC Berkeley and Princeton University. He received the B.S., M.S. and Ph.D. degrees in Electrical and Computer Engineering from Duke University in 2004, 2007 and 2010. His research is in the area of statistical machine learning and focuses on probabilistic modeling and inference techniques, Bayesian nonparametric methods, and text and image processing.
\end{IEEEbiography}
\vspace{-30pt}

\begin{IEEEbiography}[{\includegraphics[width=1in,height=1.25in,clip,keepaspectratio]{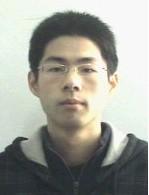}}]{Qin Lin}
is currently a graduate student in the Department of Communication Engineering at Xiamen University. His research interests includes computer vision, machine learning and data mining.
\end{IEEEbiography}
\vspace{-30pt}

\begin{IEEEbiography}[{\includegraphics[width=1in,height=1.25in,clip,keepaspectratio]{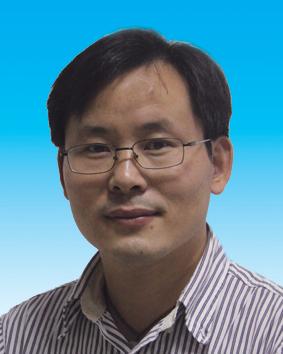}}]{Xinghao Ding}
was born in Hefei, China in 1977. He received the B.S. and Ph.D degrees from the Department of Precision Instruments at Hefei University of Technology in Hefei, China in 1998 and 2003.

From September 2009 to March 2011, he was a postdoctoral researcher in the Department of Electrical and Computer Engineering at Duke University in Durham, NC. Since 2011 he has been a Professor in the School of Information Science and Engineering at Xiamen University. His main research interests include image processing, sparse signal representation, and machine learning. 
\end{IEEEbiography}
\vspace{-30pt}

\begin{IEEEbiography}[{\includegraphics[width=1in,height=1.25in,clip,keepaspectratio]{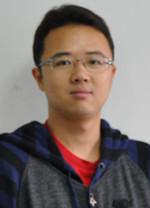}}]{Xueyang Fu}
is currently a graduate student in the Department of Communication Engineering at Xiamen University. His research interests include image processing, sparse representation and machine learning.
\end{IEEEbiography}
\vspace{-30pt}

\begin{IEEEbiography}[{\includegraphics[width=1in,height=1.25in,clip,keepaspectratio]{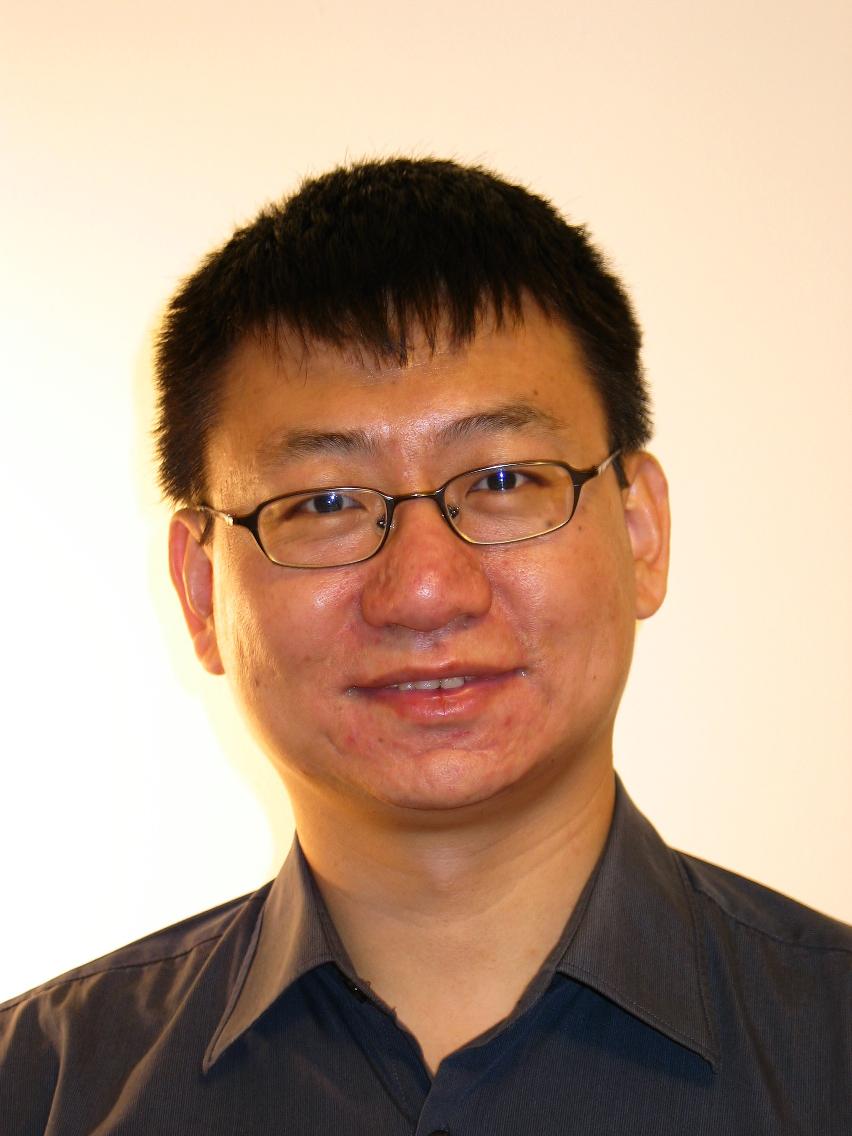}}]{Xiao-Ping Zhang}
(M'97, SM'02) received B.S. and Ph.D. degrees from Tsinghua University, in 1992 and 1996, respectively, both in Electronic Engineering. He holds an MBA in Finance, Economics and Entrepreneurship with Honors from the University of Chicago Booth School of Business, Chicago, IL.
 
Since Fall 2000, he has been with the Department of Electrical and Computer Engineering, Ryerson University, where he is now Professor, Director of Communication and Signal Processing Applications Laboratory (CASPAL). He has served as Program Director of Graduate Studies. He is cross appointed to the Finance Department at the Ted Rogers School of Management at Ryerson University. Prior to joining Ryerson, he was a Senior DSP Engineer at SAM Technology, Inc., San Francisco, and a consultant at San Francisco Brain Research Institute. He held research and teaching positions at the Communication Research Laboratory, McMaster University, and worked as a postdoctoral fellow at the Beckman Institute, the University of Illinois at Urbana-Champaign, and the University of Texas, San Antonio. His research interests include statistical signal processing, multimedia content analysis, sensor networks and electronic systems, computational intelligence, and applications in bioinformatics, finance, and marketing. He is a 
frequent consultant for biotech companies and investment firms. He is cofounder and CEO for EidoSearch, an Ontario based company offering a content-based search and analysis engine for financial data.

Dr. Zhang is a registered Professional Engineer in Ontario, Canada, a Senior Member of IEEE and a member of Beta Gamma Sigma Honor Society. He is the general chair for MMSP'15, publicity chair for ICME'06 and program chair for ICIC'05 and ICIC'10. He served as guest editor for Multimedia Tools and Applications, and the International Journal of Semantic Computing. He is a tutorial speaker in ACMMM2011, ISCAS2013, ICIP2013 and ICASSP2014. He is currently an Associate Editor for IEEE Transactions on Signal Processing, IEEE Transactions on Multimedia, IEEE Signal Processing letters and for Journal of Multimedia.
\end{IEEEbiography}

\end{document}